\pdfoutput=1

\documentclass[11pt]{article}

\usepackage[preprint]{acl}

\usepackage{times}
\usepackage{latexsym}

\usepackage[T1]{fontenc}

\usepackage[utf8]{inputenc}

\usepackage{microtype}

\usepackage{inconsolata}

\usepackage{graphicx}

\usepackage{microtype}
\usepackage{graphicx}
\usepackage{subfigure}
\usepackage{booktabs} 
\usepackage{array,multirow,graphicx}
\usepackage{amsmath}
\usepackage{amsfonts}
\usepackage{amssymb}
\usepackage{listings}
\usepackage{comment}
\usepackage{tcolorbox}
\tcbuselibrary{most}
\usepackage{float}

\usepackage{enumitem}
\setlist{nolistsep}

\newcounter{experimentcounter}  
\renewcommand{\theexperimentcounter}{\Roman{experimentcounter}}  

\newcommand{\myexperimentlabel}[1]{\refstepcounter{experimentcounter}Experiment \theexperimentcounter\label{#1}}    

\newtcolorbox{longfbox}[2][]{colframe=black, colback=white, sharp corners, 
  enhanced, title=#2, fonttitle=\bfseries, #1}

\title{Understanding Silent Data Corruption in LLM Training}

\author{%
  Jeffrey Ma$^{1}$
  \quad Hengzhi Pei$^2$
  \quad Leonard Lausen$^2$
  \quad \textbf{George Karypis}$^2$
  \\
  $^1$Department of Computer Science, Harvard University
  \quad $^2$Amazon Web Services
  \\
  \texttt{jeffreyma@g.harvard.edu} 
  \\
  \texttt{\{philepei,lausen,gkarypis\}@amazon.com} 
}

\begin{document}
\maketitle

\begin{abstract}
As the scale of training large language models (LLMs) increases, one emergent failure is silent data corruption (SDC), where hardware produces incorrect computations without explicit failure signals. In this work, we are the first to investigate the impact of real-world SDCs on LLM training by comparing model training between healthy production nodes and unhealthy nodes exhibiting SDCs. With the help from a cloud computing platform, we access the unhealthy nodes that were swept out from production by automated fleet management. Using deterministic execution via XLA compiler and our proposed synchronization mechanisms, we isolate and analyze the impact of SDC errors on these nodes at three levels: at each submodule computation, at a single optimizer step, and at a training period. Our results reveal that the impact of SDCs on computation varies on different unhealthy nodes. Although in most cases the perturbations from SDCs on submodule computation and gradients are relatively small, SDCs can lead models to converge to different optima with different weights and even cause spikes in the training loss. Our analysis sheds light on further understanding and mitigating the impact of SDCs.
\end{abstract}

\section{Introduction}

Large language models (LLMs) have demonstrated remarkable advancements in different tasks. To further explore the potentials of LLMs, recent efforts have been put into scaling up the model size to hundred-billions or even trillions of parameters. As a result, training such large models requires extensive computational resources over a long training period. For example, Llama 3 405B was trained on 16K H100 GPUs \cite{dubey2024llama3herdmodels}.

However, as training scale increases, the likelihood of hardware failures also increases. Silent data corruption (SDC) is an emerging error that causes impacted hardware to inadvertently output wrong calculation results silently without any user indication \cite{dixit2021silentdatacorruptionsscale}. Meta reported 6 unplanned job interruptions were attributed to SDC during a 54-day pre-training snapshot \cite{dubey2024llama3herdmodels} and Google estimated an SDC event occurs every week or two during Gemini training \cite{geminiteam2024geminifamilyhighlycapable}. 
In practice, SDCs observed during large-scale training usually result from latent hardware defects that cause corruption only under certain conditions or after sufficient stress over the hardware's lifetime. Once a machine begins to be affected by SDCs, it pollutes training outputs and can impact the model optimization trajectory \cite{hepermanenthardwarefailures}.
Although many works studied the effect of SDCs in large-scale CPU systems \cite{dixit2021silentdatacorruptionsscale, wang2023understandingsdc}, autonomous systems \cite{wan2022analyzing, yushunsdcrobotics2024} and deep learning accelerators \cite{zhang2018analyzing, liunderstandingerrorpropagation, rech2022reliability}, no public work has characterized the impact of real-world SDCs on LLM training in detail. 

In this work, we are the first to investigate the impact of real-world SDCs on LLM training. 
We work with a cloud-computing platform to gain access to unhealthy nodes that failed production fleet management testing due to SDCs.
While unhealthy hardware is generally excluded from production workloads, latent defects and hardware failures can turn previously healthy hardware unhealthy, emphasizing the importance of our investigation.

By leveraging deterministic execution from the XLA compiler and adopting the same training setup, we can compare the results from unhealthy nodes and healthy nodes to characterize the impact of SDCs during training.
We break down our investigation into three levels: (1) the impact on submodule computation; (2) the impact on the gradients of model weights at a single optimizer step; and (3) the impact on the model quality over a training period. Since SDC error can accumulate, we isolate the impact of SDCs at different levels by overwriting the intermediate results computed on the unhealthy node with results from the healthy node. Specifically, we design a computation synchronization mechanism to ensure the input of every submodule is the same on healthy nodes and on unhealthy nodes for (1), and a parameter synchronization mechanism to ensure the model weights are the same before each optimizer step for (2).

We conduct quantitative comparisons with the computations on healthy node at different levels. To investigate the impact of SDCs on submodule computation, we check the forward and backward computation of the self-attention and feed-forward network (FFN) across unhealthy nodes. To investigate the impact at each optimizer step, we examine the difference in gradients. To investigate the accumulated impact on the model quality, we track the loss and parameter difference during pretraining and also examine the finetuning performance on unhealthy nodes. 

Our empirical results show that SDCs do not occur uniformly during training and exhibit different patterns on different unhealthy nodes. We find that SDCs can cause certain values in the submodule computation results to differ by large factors, while the average mismatch frequency is generally low. Furthermore, the noise to gradients caused by SDC error within an optimizer step is small relative to the true gradient norm. For the accumulated impact over training steps, although the pretraining loss remains similar, SDCs can cause model parameters to drift away from ground-truth weights, which indicates that models on different nodes converge to different optima. Meanwhile, although the models fine-tuned on most unhealthy nodes have similar performance compared to the models fine-tuned on the healthy node, loss spikes do occur during fine-tuning on some unhealthy nodes, which can fully corrupt the model weights in some cases.

In summary, our contributions are:
\begin{itemize}
    \item We are the first to investigate the impact of read-world SDCs on LLM training in detail by obtaining access to realistic unhealthy nodes flagged by the production fleet management.
    \item By pairing unhealthy nodes with healthy nodes and introducing synchronization mechanisms, we design experiments to precisely isolate the impact of SDCs at different levels.
    \item We reveal the characteristics of SDCs at various levels of model training empirically and further provide insightful analysis which sheds light on the future work on understanding and mitigating the impact of SDCs.
\end{itemize}

\section{Background} \label{sec:background}

\subsection{Silent Data Corruption Errors} \label{sec:sdc_background} 
Silent Data Corruption (SDC) errors are incorrect computation that silently occur during normal usage \cite{papadimitriousilentdatacorruptions2023}. Generally, SDCs can arise from hardware faults \cite{dixit2021silentdatacorruptionsscale, hochschild2021cores}, environmental factors like radiation \cite{ziegler1996terrestrial, mukherjee2005soft, baumann2005radiation}, or software bugs \cite{lou2022demystifying}. 
With growth of large scale distributed systems, SDC is observed to be caused by device-specific hardware defects which show errors at certain utilization levels or temperatures \cite{ dixit2021silentdatacorruptionsscale, hochschild2021cores, wang2023understandingsdc}. 

Our work lies in the broader area of understanding the effect of SDCs on deep learning and we leave a detailed discussion for the related work in Appendix \ref{appendix:sdc_related_work}.
There are two limitations in this area. 
First, fault-injection methods are commonly used for evaluation. Although SDC can be simulated at the hardware level \cite{rech2022reliability, liunderstandingerrorpropagation} or the software level \cite{he2020fidelity, agarwal2022lltfi}, simulated SDCs could be different from those observed in production. 
Second, most works study the impact of SDCs on model inference \cite{liunderstandingerrorpropagation, agarwal2023resilience, ma2023evaluatingenhancingrobustnessdeep} while few study the impact of SDCs on model training dynamics. Although some empirical findings of SDC like NaN (Not-a-Number) issues and degradation during training are reported \cite{adept-sdc, he2023understanding}, there is no work that attempts to characterize and break down the impacts of real-world SDCs especially on large language model training.

\subsection{Large Language Model Training}
In this work, we consider a Large Language Model (LLM) as a transformer decoder \cite{vaswani2017attentionneed}. Training an LLM at scale requires a combination of data parallelism and model parallelism. Tensor parallelism (TP) is a widely used model parallelism approach that partitions each individual layer of a model across accelerators \cite{shoeybi2020megatronlmtrainingmultibillionparameter}. For large-scale LLM training, tensor-parallel size is usually set as the number of accelerators within a node to leverage high bandwidth intra-node communication \cite{narayanan2021efficient}.

Given that hardware failure is usually flagged at the node level and majority of training failures are caused by one node \cite{wang2023gemini}, we focus on the impact of SDCs on the computation of a single node and use tensor parallelism only.

\section{Methodology} \label{sec:methodology}

In this section, we discuss our methodology for investigating the effect of real-world SDCs on LLM training. We first describe a high-level mechanism used to collect the unhealthy nodes for our experiments. Then, we break down the investigation of the SDC impact into three granularities and ask three key research questions (RQs). To answer these RQs, we further propose two synchronization mechanisms to isolate the impact of SDC.

\subsection{Hardware Collection in the Real World} 

\begin{figure}[t]
    \centering
    \includegraphics[width=0.95\linewidth]{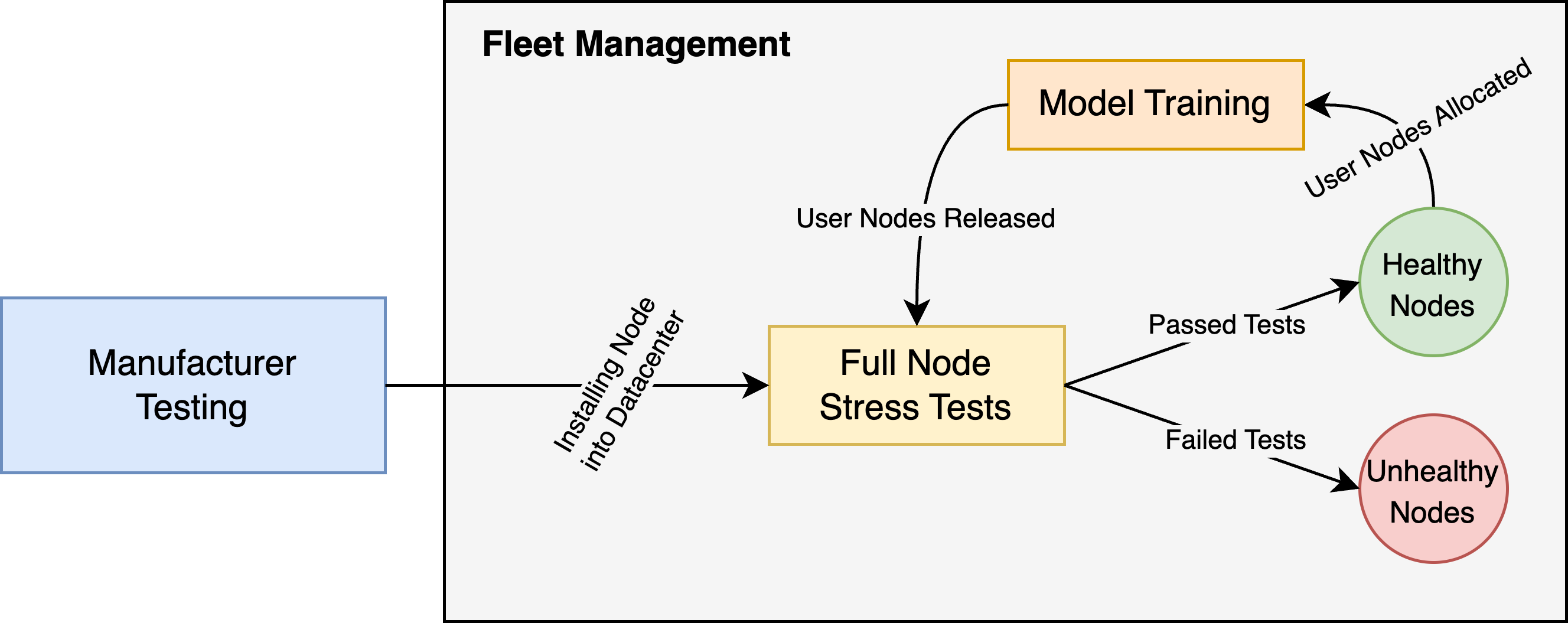}
    \vskip -0.05in
    \caption{Illustration of fleet management flow, where nodes are vetted through several rounds of testing at different granularities.}
    \vskip -0.2in
    \label{fig:fleet_management}
\end{figure}

To study the effect of real-world SDCs, we identify \emph{nodes that failed production margin stress tests and thus were not allowed into production availability}. Figure \ref{fig:fleet_management} shows a high-level production fleet management flow for identifying failing nodes. Before entering this flow, components such as machine learning accelerators go through stages of testing at the manufacturer and after system assembly.

\label{sec:sdc_fleet_management}

Once a node is installed in a datacenter, additional stress tests are triggered to identify any compound hardware issue.
These tests include full-node communication collective stress tests, compute unit stress tests, and a small LLM training run where training outputs are compared with pre-computed golden truth values.
With deterministic workload execution, the difference from ground truth values or other non-determinism indicates SDC on the node.
To guard against hardware degradation over time, tests are also run when a node is reclaimed from customer usage either due to customer workloads ending without any indication of error or because the customer returned the hardware after receiving a signal indicating a hardware health issue from the cloud provider. 

Using this flow, we define two types of nodes:
\begin{itemize}
    \item \underline{\emph{Unhealthy nodes}} are nodes that fail the fleet management tests due to exhibiting SDC. 
    \item \underline{\emph{Healthy nodes}} are nodes from production that have passed the aforementioned tests. 
\end{itemize}
Each category contains fifteen (15) nodes. We have confirmed that \emph{all healthy nodes in our experiments will output the same result for the same computation} and we denote them as the healthy node for simplicity. However, unhealthy nodes can exhibit different symptoms for SDC and we assign each unhealthy node a unique identifier, namely Node 1 to Node 15. 

Unhealthy nodes flagged by fleet management are meaningful for studying real-world SDCs. Given that fleet management can only be run when the nodes are not used by customers, during multi-month periods of a large-scale LLM training run, healthy nodes that were originally healthy can degrade and produce SDCs affecting training before fleet management can isolate them.
We confirm that some of the unhealthy nodes in our experiments were initially healthy and began to fail the pre-checks after being used in training.

\subsection{Key Research Questions} \label{sec:research_questions}
To better understand the impact of SDCs, we break down our investigation into three levels and ask three main research questions (RQs):

\underline{\textbf{RQ1}}: \emph{What is the impact of SDCs on Transformer submodule computation results?}

\underline{\textbf{RQ2}}: \emph{What is the impact of SDCs on the gradients of model weights at a single optimizer step?} 

\underline{\textbf{RQ3}}: \emph{What is the accumulated impact of SDCs on the model quality over training steps?}

Investigating \textbf{RQ1} helps us understand the frequency and severity of SDCs, critical for designing a solution to detect SDC at the submodule level. Investigating \textbf{RQ2} and \textbf{RQ3} gives us more insight into understanding optimization dynamics when SDCs occur.
Most importantly, these research questions help prioritize future directions of detecting, mitigating, and recovering from SDC by providing concrete real-world SDC characteristics.

To compare SDC-induced incorrect computations with ground truth, we pair each unhealthy node with a healthy node and train identical models simultaneously on each node with exactly the same 
training setup. 
We employ the XLA compiler \cite{xlapaper} to ensure fully deterministic instruction ordering to isolate away non-SDC sources of non-determinism like floating point error. 
In other words, we confirm that \emph{the computation results are exactly the same on any two healthy nodes with the same compiler and the difference in computation results can be entirely attributed to SDC.}

Since SDC error can accumulate over computation, we need to correct the error on unhealthy nodes with the ground-truth results on the healthy node to isolate the impact of SDCs at different levels. Specifically, we design two synchronization mechanisms for \textbf{RQ1} and \textbf{RQ2}.

\begin{figure}[t]
    \centering
    \includegraphics[width=\linewidth]{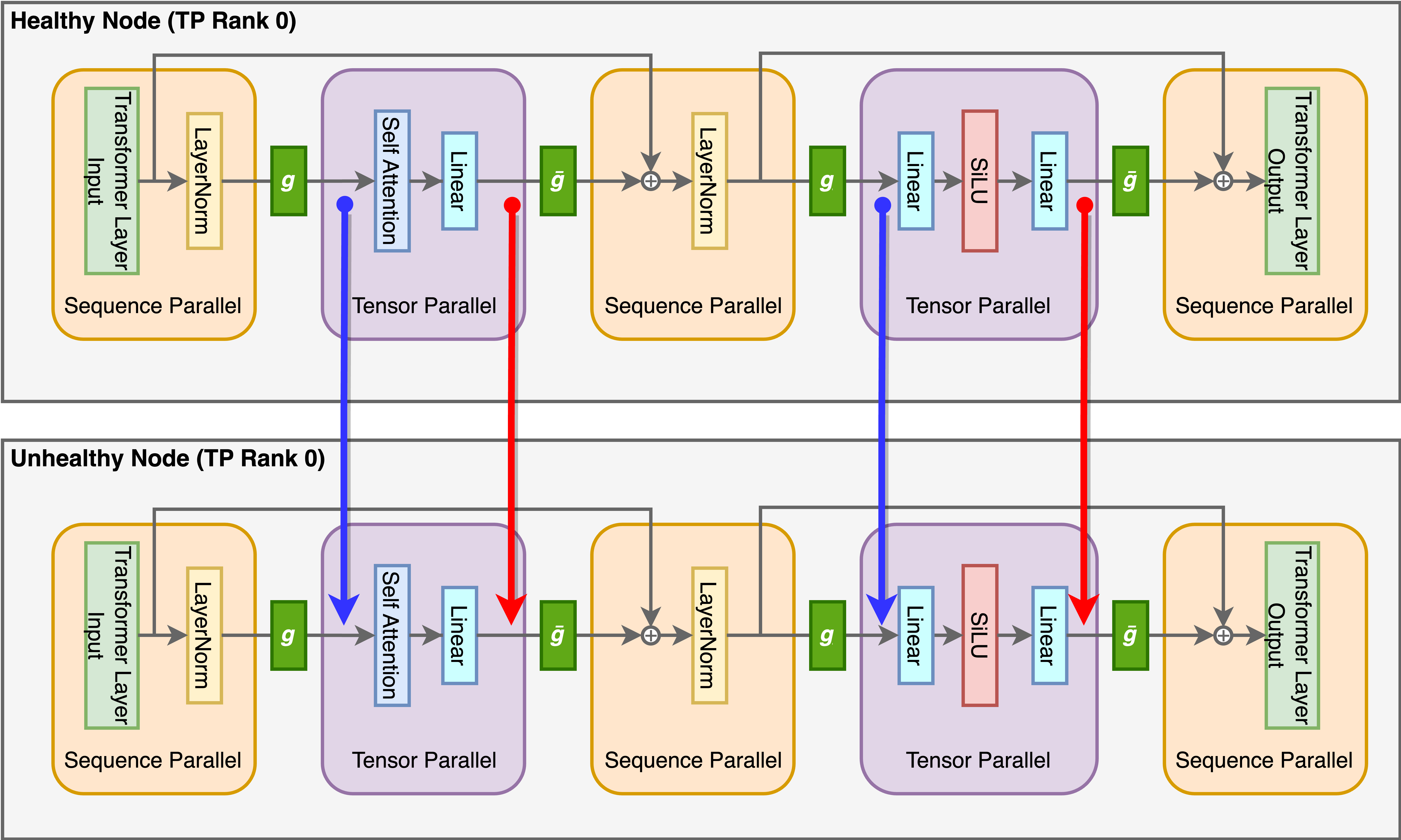}
    \vskip -0.05in
    \caption{Illustration of our lock-step parallelism works in a Transformer decoder layer. The arrows indicate intermediate tensors corrected by communicating values from the healthy to the unhealthy node (red in forwards pass, blue in backwards pass). In forward pass, $g$ is an all-gather and $\bar{g}$ is a reduce-scatter, while in the backwards pass $g$ is an reduce-scatter and $\bar{g}$ is an all-gather.}
    \vskip -0.2in
    \label{fig:primitive_investigation}
\end{figure}

\textbf{Computation Synchronization.}  To isolate the impact of SDCs at submodule level (\textbf{RQ1}), we set up a novel communication mesh called \textbf{lock-step parallelism} shown in Figure \ref{fig:primitive_investigation}. 
For each pair of same TP ranks on the unhealthy and healthy node, in the forward pass, we communicate and compare the output values after the forward computation of each submodule (self-attention or FFN) before the reduce-scatter of sequence parallelism, which avoids incorrect values on certain ranks spreading to tensor shards on other ranks after communication. Then, we overwrite the values on the unhealthy node with those from the healthy node (red arrows) to prevent SDC error from accumulating to the next submodule. Likewise, in the backward pass, we compare the input gradient after the backward computation of each submodule before the reduce-scatter (blue arrow) and overwrite the gradient values to prevent error from accumulating through backpropagation. More details can be found in Appendix \ref{appendix:primivite_investigation}.

\textbf{Parameter Synchronization.} To isolate the impact of SDCs at a single training step (\textbf{RQ2}), we do parameter synchronization at the end of each training step.
After taking an optimizer step, we broadcast the updated model parameters from the healthy node to overwrite the parameters on the unhealthy node. In this way, both nodes start from the same parameters for the next optimizer step.

For our synchronization mechanisms, we assume that \emph{SDCs do not occur when communicating tensors between the unhealthy and healthy node}. 
First, all nodes used in our experiments consistently passed stress tests for communication collectives. Second, the communication primitives used in our synchronization mechanism do not involve any compute unit. Finally, parity checks or error correcting codes are utilized for communication across the network. 
To avoid SDCs occurring during the arithmetic of tensor comparisons between healthy and unhealthy nodes, we always compute the comparisons on the healthy node.

\section{Experiments for RQ1}
\label{sec:primitive}
\underline{\textbf{RQ1}}: \emph{What is the impact of SDCs on Transformer submodule computation outputs?}

In this section, we first introduce our experiment setups for RQ1 and analyze experimental results to understand the impact of SDCs on submodule computation. We follow the same structure for RQ2 in  Section \ref{sec:sdc_single_optimizer_step} and RQ3 in Section \ref{sec:multiple_training_steps}.

\subsection{Experiment Setups}
\label{methods:primitive_impact}
We focus on four kinds of Transformer submodule computation, namely the forward computation of a self-attention module (FWD/ATTN) and an FFN module (FWD/FFN), and the backward computation of a self-attention module (BWD/ATTN) and an FFN module (BWD/FFN). We train two models on each pair of the unhealthy node and the healthy node simultaneously and use the \textbf{computation synchronization} mechanism discussed in Section \ref{sec:research_questions}. We use a decoder-only Transformer architecture similar to the Llama3-8B configuration \cite{dubey2024llama3herdmodels} with $D=16$ decoder layers and hidden state size of $H=4096$ and use the tensor parallelism to fit a model within a node. More details can be found in Appendix \ref{appendix:primivite_investigation}.

For a submodule computation $f$ in the model, we define $f'_{i}(x_{t,j})$ as the tensor computed on TP rank $t$ of unhealthy node $i$ at the microstep $j$ and $f(x_{t,j})$ as the corresponding output on healthy node. To quantify differences between $f'_{i}(x_{t,j})$ and $f(x_{t,j})$, we define two metrics called \emph{mismatch frequency} and \emph{mismatch severity}. We calculate the mismatch frequency for submodule $f$ on unhealthy node $i$ at the microstep $j$ as follows:
\begin{equation}
    freq^{f}_{i,j} = \frac{\sum_{t=1}^{TP}{Mis(f'_{i}(x_{t,j}), f(x_{t,j}))}}{TP \cdot MBS \cdot L\cdot H}
\end{equation}
where $Mis(y', y)$ counts the number of mismatching elements in two tensors $y$ and $y'$. We report the aggregated mismatch frequency for each submodule computation type $F$ on unhealthy node $i$ at the microstep $j$ by averaging across decoder layers:
\begin{equation}
    \label{equation:mismatch_frequency}
    freq^{F}_{i,j} = \frac{1}{D}\sum_{f \in F}{freq^{f}_{i,j}}, F=\{f^{(1)},...,f^{(D)}\}
\end{equation} 
Mismatch severity is defined as the average over non-zero values of the element-wise relative difference. Formally, we take the maximum over all TP ranks and calculate the mismatch severity for submodule $f$ on unhealthy node $i$ at microstep $j$:
\begin{equation}
    sev^{f}_{i,j} = \max_{0 \leq t < TP}{\left[
        {Avg_{\neq 0}}\left({
            \left|
                \frac{f'_{i}(x_{t,j}) - f(x_{t,j})  }{f(x_{t,j})}
            \right|
        }\right)
    \right]}
\end{equation}
where $Avg_{\neq 0}(x)$ computes the average value over only non-zero elements of a tensor $x$. We also calculate the mismatch severity for each type of submodule computation $F$ on unhealthy node $i$ at the microstep $j$ by taking the maximum across decoder layers as follows:
\begin{equation}
    \label{equation:mismatch_severity_max_decoder}
    sev^{F}_{i,j} = \max_{f \in F}{sev^{f}_{i,j}}, F=\{f^{(1)},...,f^{(D)}\}
\end{equation}

\subsection{Results} \label{sec:submodule_outputs_results}
Table \ref{fig:primitive_frequency_summarized} shows the mismatch frequency of submodule computation.
We observe that the impact of SDCs on submodule computation varies across different unhealthy nodes, e.g. Nodes 10 and 11 have a high mismatch frequency while Nodes 2 and 3 do not show any SDC occurrence in this setting. 

\begin{table}[t]
\begin{center}
\begin{small}
\begin{sc}
\setlength{\tabcolsep}{3pt}
\renewcommand{\arraystretch}{1.1} 
\begin{tabular}{crrrr}
\toprule
Node ID & fwd/attn & fwd/ffn & bwd/attn & bwd/ffn  \\ \midrule
Node 1  & 1.55$e$-5  & 5.06$e$-7  & 1.56$e$-04  & 2.81$e$-6   \\
Node 4  & 3.79$e$-9  & 9.20$e$-11  & 2.99$e$-9  & 2.80$e$-11  \\
Node 5  & 0  & 0  & 1.49$e$-15  & 1.25$e$-12  \\
Node 6  & 1.71$e$-9  & 1.64$e$-11  & 1.49$e$-9  & 6.02$e$-11   \\
Node 7  & 2.13$e$-6  & 1.18$e$-7  & 4.31$e$-6  & 6.73$e$-8   \\
Node 8  & 3.21$e$-9  & 1.99$e$-11   & 1.01$e$-7  & 2.21$e$-9   \\
Node 9  & 1.10$e$-5  & 5.05$e$-7   & 4.33$e$-6  & 3.86$e$-8  \\
Node 10 & 4.78$e$-3  & 1.03$e$-3    & 1.92$e$-3  & 7.98$e$-5   \\
Node 11 & 2.89$e$-2  & 2.25$e$-3  & 6.71$e$-3  & 1.08$e$-4   \\
Node 13 & 0  & 0   & 0  & 1.21$e$-10  \\
Node 14 & 6.48$e$-11  & 0   & 4.91$e$-10  & 2.99$e$-9   \\
Node 15 & 0  & 0   & 0  & 7.39$e$-15  \\
\bottomrule
\end{tabular}
\end{sc}
\end{small}
\end{center}
\vskip -0.1in
\caption{Average mismatch frequency over microsteps for Transformer submodules. The table excludes the nodes that do not show any SDC in this setting.}
\vskip -0.1in
\label{fig:primitive_frequency_summarized}
\end{table}

\begin{figure}[t!]
    \centering
    \includegraphics[width=0.85\linewidth]{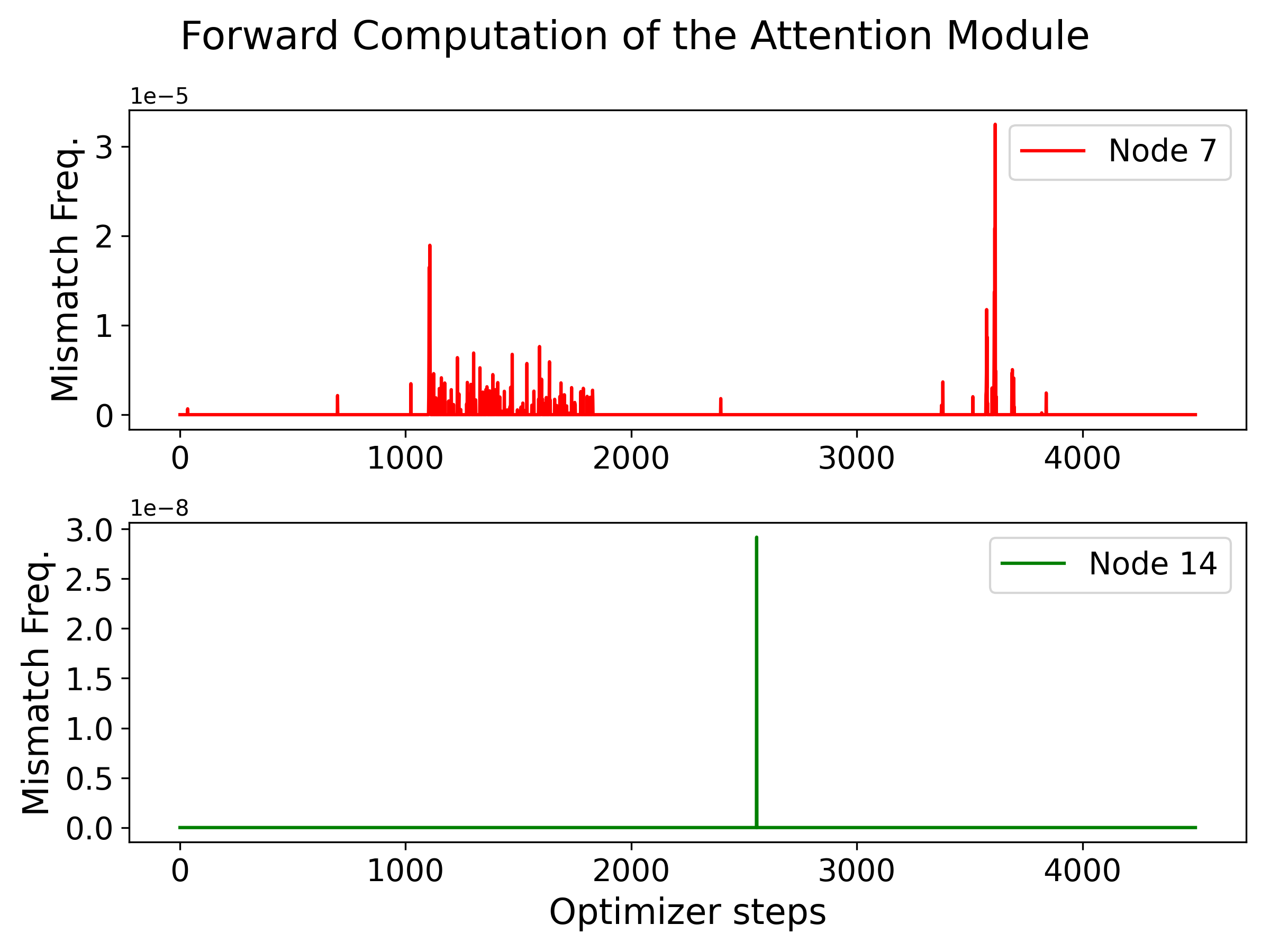}
    \vskip -0.15in
    \caption{Non-uniform spikes of mismatch frequency in the forward computation of the attention module over time on Node 7, 14.}
    \label{fig:attention_forward_mismatch_frequency_over_time}
\vskip -0.15in
\end{figure}

\begin{figure}[t]
    \vskip 0.1in
    \centering
    \includegraphics[width=0.85\linewidth]{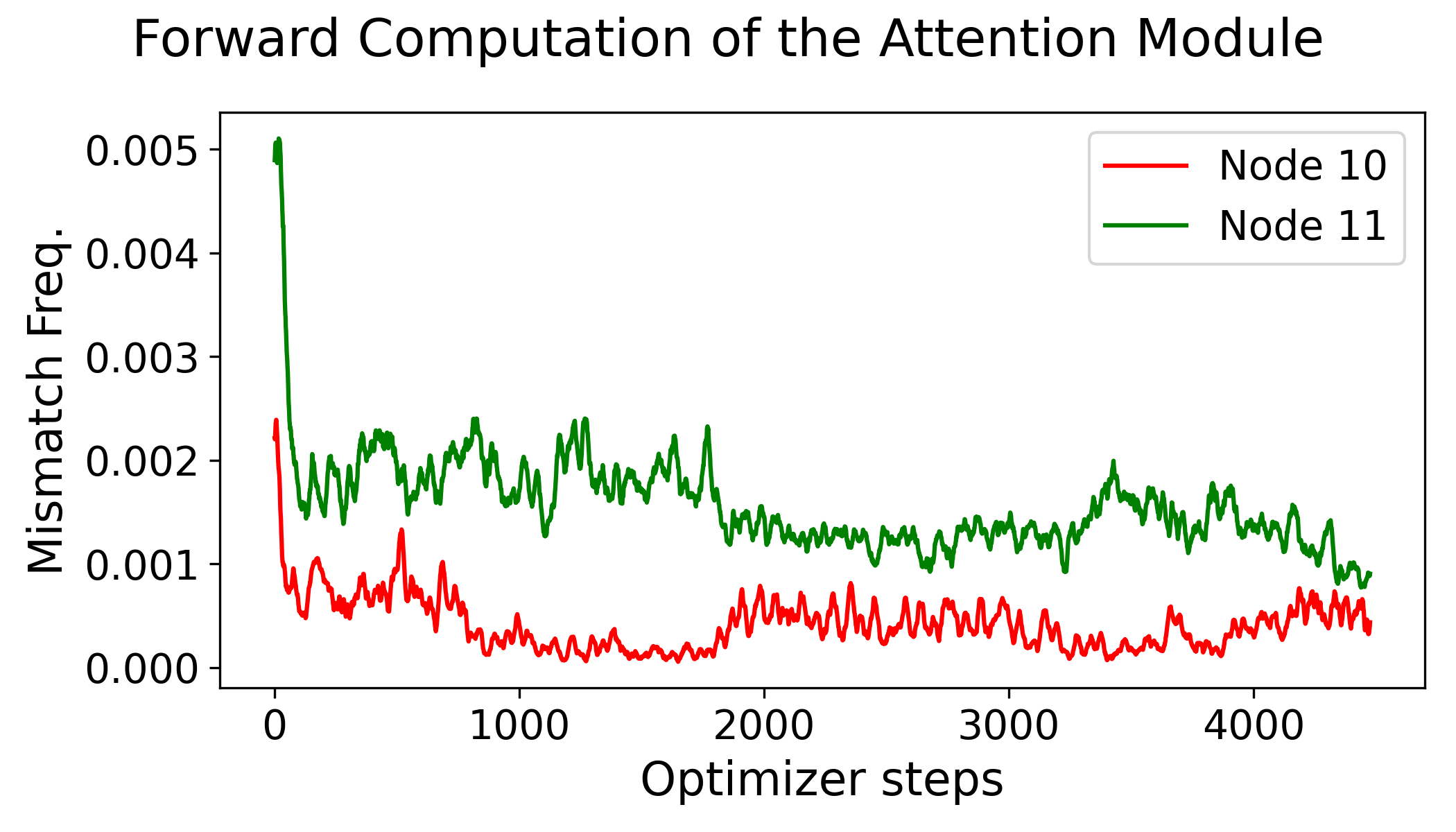}
    \vskip -0.1in
    \caption{High SDC occurrence with large initial spikes in smoothed mismatch frequency for the forward computation of the attention module on Node 10, 11.}
    \label{fig:primitive_spike_at_beginning}
    \vskip -0.1in
\end{figure}

We further find that SDCs do not occur uniformly over time: mismatch frequency often has a large variance across steps. Figure \ref{fig:attention_forward_mismatch_frequency_over_time} shows the mismatch frequency in the forward computation of the attention module on Nodes 7 and 14. We find that spikes of mismatch frequency sometimes occur, while during the majority of training time, no mismatch occurs. In Figure \ref{fig:primitive_spike_at_beginning}, we observe a high peak of mismatch frequency at the first few steps on Nodes 10 and 11, likely due to higher overall system usage when initializing the training run. The non-uniform occurrence of SDC during training suggests that SDCs might be caused by broader, compound system-level factors.

Table \ref{fig:primitive_severity_summarized}  shows the maximum mismatch severity over optimizer steps for submodule computation on different unhealthy nodes. We find that SDCs cause certain tensor values in the computation to differ by large factors. For example, on Node 9, the mismatch severity exceeds $100$, which means SDCs can cause degraded TP ranks to have very different computation results on certain microsteps.

\begin{table}[t]
\begin{center}
\begin{small}
\begin{sc}
\setlength{\tabcolsep}{3pt}
\renewcommand{\arraystretch}{1.1} 
\begin{tabular}{crrrr}
\toprule
Node ID & fwd/attn & fwd/ffn & bwd/attn & bwd/ffn \\
\midrule
Node 1 & 99 & 312 & 7392 & 3.88 \\
Node 4 & 2.95 & 1.46 & 2.39 & 5.38 \\
Node 5 & 0 & 0 & 0.0047 & 0.0908 \\
Node 6 & 1.01 & 1 & 0.162 & 0.0776 \\
Node 7 & 43.2 & 88.5 & 32.3 & 0.297 \\ 
Node 8 & 13.5 & 5.69 & 6.41 & 1.59 \\ 
Node 9 & 119 & 200 & 3.79$e$+11 & 9.63$e$+12 \\ 
Node 10 & 1120 & 262 & 12.75 & 0.648 \\
 Node 11 & 318 & 976 & 2208 & 7680 \\ 
Node 13 & 0 & 0 & 0 & 0.316 \\ 
Node 14 & 1.12 & 0 & 3.80 & 0.380 \\ 
Node 15 & 0 & 0 & 0 & 0.0176 \\ \bottomrule
\end{tabular}

\end{sc}
\end{small}
\end{center}
\vskip -0.1in
\caption{Maximum mismatch severity over microsteps for each unhealthy node. The table excludes the nodes that do not show any SDC in this setting.}
\vskip -0.1in
\label{fig:primitive_severity_summarized}

\end{table}

\begin{figure*}[ht]
    \centering
    \vskip -0.12in
    \begin{minipage}[c]{0.69\linewidth}
        \centering
        \includegraphics[width=\linewidth]{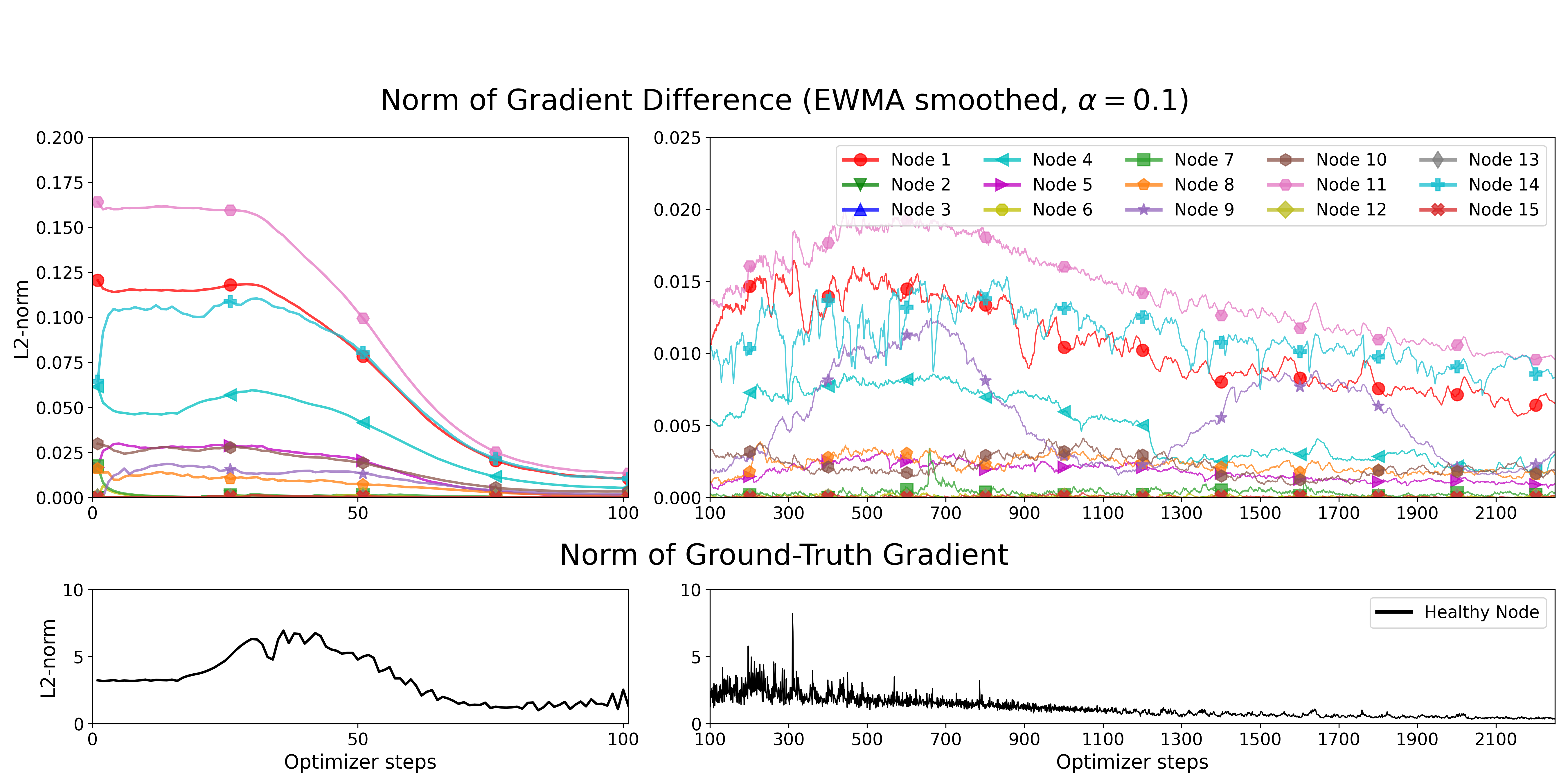}
    \end{minipage}
    \hfill
    \begin{minipage}[c]{0.3\linewidth} 
        \centering
        \vskip 0.2in
        \begin{tiny}
            \begin{sc}
                \begin{tabular}{cc} 
                    \toprule
                    Node ID & WCNTS Ratio \\
                    \midrule
                    1 & 0.037 \\
                    2 & 0.002 \\
                    3 & 1.34$e$-15 \\
                    4 & 0.019 \\
                    5 & 0.011 \\
                    6 & 0.004 \\
                    7 & 0.008 \\
                    8 & 0.006 \\
                    9 & 0.017 \\
                    10 & 0.010 \\
                    11 & 0.051 \\
                    12 & 0.002 \\
                    13 & 0.004 \\
                    14 & 0.037 \\
                    15 & 4.16$e$-4 \\
                    \bottomrule
                \end{tabular}
            \end{sc}
        \end{tiny}

    \end{minipage}
    \vskip -0.1in
    \caption{$L_2$-norm of the gradient difference and the ground-truth gradients over steps. The left table shows Worst Case Noise-to-Signal (WCNTS) ratios for unhealthy nodes.}
    \label{fig:fix_weight_only}
    \vskip -0.1in
\end{figure*}

\section{Experiments for RQ2} \label{sec:sdc_single_optimizer_step}
\underline{\textbf{RQ2}}: \emph{What is the impact of SDCs on the gradients of model weights at a single optimizer step?}

\subsection{Experiment Setups}
\label{methods:differing_gradients}
We train same models on each pair of unhealthy and healthy nodes simultaneously and use the \textbf{parameter synchronization} mechanism discussed in Section \ref{sec:research_questions}. 
After the forward and backward pass are finished  at step $j$, we compute the $L_2$ norm of elementwise difference between the gradients of model weights on the $i$-th unhealthy node $g'_{i,j}$ and the ground-truth gradients on the healthy node $g_{j}$. After taking an optimizer step, we use parameter synchronization to overwrite the model parameters on the unhealthy node. 
We also report the \emph{worst case noise-to-signal (WCNTS) ratio} to measure how significant SDC-induced noise to gradients is:
\begin{equation}
    \label{equation:worst_case_noise_to_signal}
    WCNTS_i = \max_{j}{\frac{\|g'_{i,j} - g_{j} \|_2}{\|g_{j}\|_2}}
\end{equation}
Using the same decoder block architecture as in Section \ref{sec:primitive}, we train a $32$-layer Transformer decoder with hidden state size as $H=4096$. More details can be found in Appendix \ref{appendix:model_training}.

\subsection{Results}
\label{results:model_gradient_norm_diff}

Figure \ref{fig:fix_weight_only} shows the $L_2$ norm of gradient difference and the WCNTS ratio for the gradients over optimizer steps across unhealthy nodes. We observe that gradients computed on unhealthy nodes deviate minimally from those computed on the healthy node. Although the absolute value of $L_2$ norm of the gradient difference is large before the $100$-th step, it is still relatively small compared to the $L_2$ norm of the ground-truth gradients and continues to decrease as the norm of the ground-truth gradients decreases.
In the worst case on Node 11, the $L_2$ norm of gradient difference is $5.1\%$ of that of the ground-truth gradients, showing that the SDC-induced noise in the gradients is relatively small relative to the ground-truth gradients. 

\section{Experiments for RQ3} \label{sec:multiple_training_steps}

\underline{\textbf{RQ3}}: \emph{What is the accumulated impact of SDCs on the model quality over multiple training steps?}

\subsection{Experiment Setups}
To provide a better understanding of how different the learned representations and model decision boundaries from unhealthy nodes would be, we design experiments for both model pre-training from scratch and fine-tuning of a pre-trained model.

\textbf{\myexperimentlabel{experiment:parameter_drift}}: \emph{How different is the learned model under accumulated SDC error from the ground truth during model pre-training?} 
We pretrain same models on each pair of unhealthy node and healthy node simultaneously. We follow the same experiment setting in Section \ref{methods:differing_gradients} except we \emph{do not use parameter synchronization mechanism}. To observe how SDCs impact the learned models during training, we report the training loss and the \emph{parameter difference} which is $L_2$ norm of the element-wise difference between model parameters on healthy and unhealthy node.

\textbf{\myexperimentlabel{experiment:finetuning}}: \emph{For downstream tasks, how would SDCs affect model finetuning?} 
We want to further understand how model quality is affected by SDCs when the model is fine-tuned on a downstream task. 
In this experiment, we fine-tune \verb|Mistral-7B-v0.3| \cite{jiang2023mistral7b} on six multiple-choice question answering tasks (CosmosQA \cite{cosmos-qa}; MathQA \cite{mathqadataset}; ScienceQA \cite{lu2022learn}; OpenbookQA \cite{OpenBookQA2018}; BoolQ \cite{clark2019boolqexploringsurprisingdifficulty}; and RACE \cite{raceLai2017large}) by instruction tuning \cite{weifinetuned}. 
For each task, we use a fixed random seed for shuffling the training dataset and evaluate the test accuracy (TA) of the models fine-tuned on the healthy node and on unhealthy nodes. Using the predictions of the model fine-tuned on the healthy node as the standard, we report the disagreement percentage (DP), which is defined as the percentage of the difference in predictions on the test set. To better understand the prediction difference, we fine-tune a model on healthy node with a different random seed as a baseline to contextualize the effect of SDC error relative to the effect of data ordering. More details can be found in Appendix \ref{appendix:finetuning}.

\subsection{Results}
\label{results:model_quality}

\begin{figure}[t]
    \centering
    \includegraphics[width=0.85\linewidth]{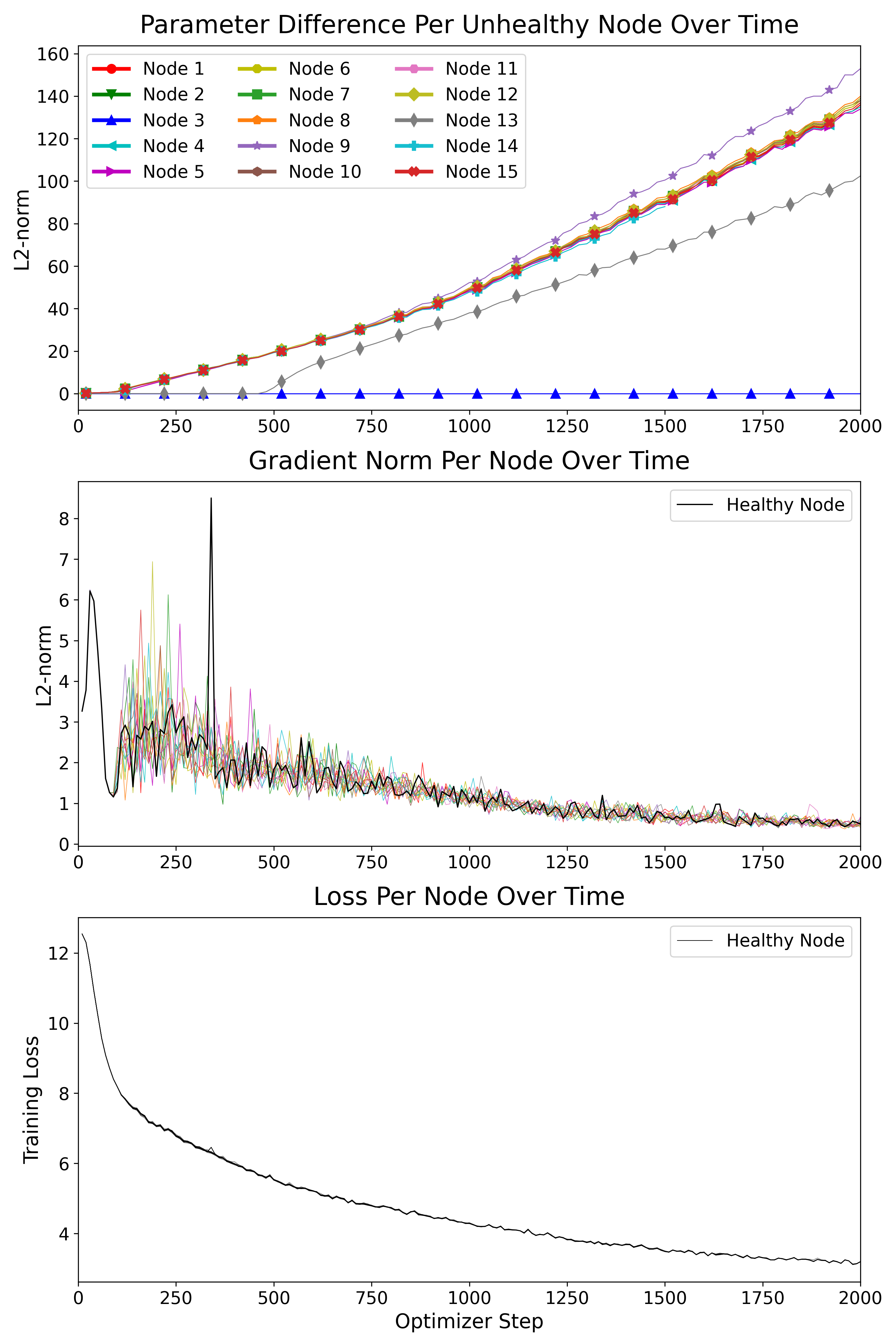}
    \vskip -0.15in
    \caption{The curves for parameter difference, gradient norms and training loss on unhealthy nodes. Note that the loss curves on all unhealthy nodes are plotted but identical to that on the healthy node.}
    \label{fig:no_synchronization}
    \vskip -0.2in
\end{figure}

\textbf{Results for Experiment \ref{experiment:parameter_drift}.} 
Figure \ref{fig:no_synchronization} shows the parameter difference, gradient norm and training loss on unhealthy nodes during pre-training. Despite training loss on each unhealthy node nearly identical to the healthy node, model weights on unhealthy nodes incrementally drift away from those on the healthy node, suggesting that SDCs are pushing models towards different local minima. 

Note that Node 13 shows no parameter difference before step 450, which indicates that no SDC occurs during this period. It is aligned with the finding in Section \ref{sec:submodule_outputs_results} that SDCs do not occur uniformly during training. After step 450, the model on Node 13 begins to quickly drift away from the ground-truth weights. We further find that the rates at which the parameter differences increase are similar on most unhealthy nodes, although the unhealthy nodes produce SDCs with different degrees of frequency and severity as shown in Section \ref{sec:submodule_outputs_results} This suggests that the rate of parameter drift is more likely to be driven by the sharp loss surface than purely by SDCs. In other words, SDCs serve as a trigger to push the optimization trajectory onto a different path through this sharp loss landscape, leading to the divergence of the parameters. 

\begin{table}[t]

\begin{center}
\begin{tiny}
\begin{sc}
\setlength{\tabcolsep}{3pt}
\begin{tabular}{cccc}
\toprule
\multirow{2}{*}{Configuration} & CosmosQA & MATH & OpenbookQA \\
 & TA (DP) & TA (DP) & TA (DP) \\
\midrule
without fine-tuning & 56.33 (44.80) & 24.02 (82.35) & 74.70 (29.30) \\
\midrule
Healthy Node & 90.79 (-) & 37.22 (-) & 83.80 (-) \\
Healthy Node (Seed=$43$) & 89.50 (6.70) & 38.83 (56.75) & 86.30 (18.70) \\
\midrule
Unhealthy Node 1 & 90.53 (5.15) & 36.78 (42.24) & 85.00 (16.80) \\
Unhealthy Node 2 & 90.79 (0.00) & 37.22 (0.00) & 83.80 (0.00) \\
Unhealthy Node 3 & 90.77 (4.96) & 34.47 (41.84) & 83.40 (16.10) \\
Unhealthy Node 4 & 90.32 (5.59) & 36.42 (37.76) & 84.30 (16.60) \\
Unhealthy Node 5 & 90.79 (0.00) & 37.19 (34.91) & 85.10 (14.10) \\
Unhealthy Node 6 & 0.00 (100.00) & 36.92 (36.82) & 84.70 (16.30) \\
Unhealthy Node 7 & 90.32 (4.99) & 37.22 (0.00) & 83.80 (0.00) \\
Unhealthy Node 8 & 89.84 (6.23) & 38.22 (36.62) & 85.20 (15.50) \\
Unhealthy Node 9 & 89.97 (5.38) & 35.78 (37.49) & 85.00 (17.40) \\
Unhealthy Node 10 & 89.97 (4.93) & 37.05 (37.05) & 84.10 (17.20) \\
Unhealthy Node 11 & 90.61 (4.54) & 36.82 (38.53) & 87.10 (15.60) \\
Unhealthy Node 12 & 90.79 (0.00) & 37.22 (0.00) & 83.80 (0.00) \\
Unhealthy Node 13 & 90.79 (0.00) & 37.22 (0.00) & 83.80 (0.00) \\
Unhealthy Node 14 & 89.74 (6.12) & 38.63 (32.29) & 85.00 (15.90) \\
Unhealthy Node 15 & 90.77 (3.27) & 38.53 (40.87) & 83.80 (0.00) \\
\bottomrule
\end{tabular}

\end{sc}
\end{tiny}
\end{center}
\vskip -0.1in
\caption{Finetuning results for three question answering tasks on different nodes. For each task, we report the test accuracy (TA) and the disagreement percentage (DP).}
\label{tab:finetuning_results_short}

\vskip -0.2in
\end{table}

\textbf{Results for Experiment \ref{experiment:finetuning}.} 
Table \ref{tab:finetuning_results_short} shows the fine-tuning results for three of the question-answering tasks on different nodes. The full results can be found in Appendix \ref{appendix:full_finetuning_results}. We find that the models fine-tuned on most unhealthy nodes are significantly better than the base model without fine-tuning and also have similar performance compared to the models fine-tuned on the healthy node. The disagreement percentage caused by SDCs on unhealthy nodes is not larger than using a different random seed for data shuffling. Aligned with the findings in Experiment \ref{experiment:parameter_drift}, this again affirms that SDCs on unhealthy nodes push the models towards different local minima.

\begin{figure}[t]
    \centering
    \includegraphics[width=0.85\linewidth]{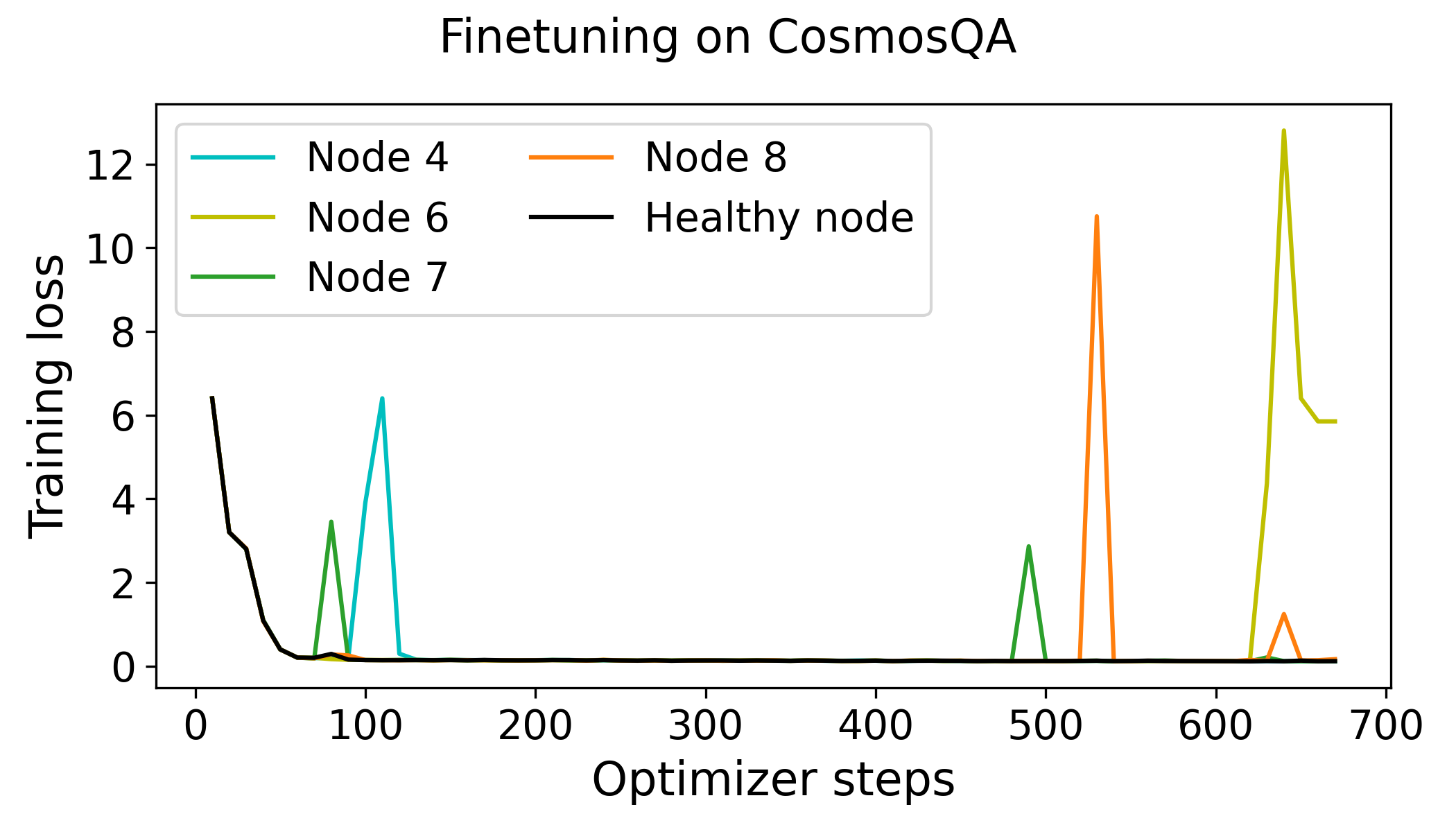}
    \vskip -0.15in
    \caption{Training loss curves on different nodes during fine-tuning on CosmosQA dataset. The loss curves for other unhealthy nodes that are identical to the healthy node are not shown in this figure.}
    \label{fig:finetuning_loss_spikes}
    \vskip -0.2in
\end{figure}

However, SDCs are not necessarily harmless to model fine-tuning. Figure \ref{fig:finetuning_loss_spikes} shows the training loss during fine-tuning on CosmosQA and we find that significant loss spikes can occur on some unhealthy nodes. On Node 4, Node 6 and Node 7, the loss spikes occur in the middle of fine-tuning while later the training is again stabilized, which makes the final models still have benign performance. However, on Node 6, the loss spike occurs near the end of fine-tuning, which leads to the resulting model having zero test accuracy on CosmosQA as shown in Table \ref{tab:finetuning_results_short}. It indicates that the loss spikes caused by SDCs pose a threat to the model quality. We also note that loss spikes do not occur in every fine-tuning task. For example, training loss curves for OpenbookQA on unhealthy nodes are all identical to that on the healthy node, similar to the situation in Experiment \ref{experiment:parameter_drift}. Therefore, we conclude that the impact of SDC on model training is closely related to the loss surface of the training task.

\section{Discussion}

\subsection{Silent Nature of SDCs in LLM Training}

Our results show that SDCs can silently occur without any clear indication from training loss. For example, we find from Section \ref{sec:submodule_outputs_results} and Section \ref{results:model_gradient_norm_diff} that SDCs on Node 10 consistently perturb the submodule computation and the gradients, but the training loss on Node 10 is still identical to that the on healthy node without any loss spike during both pre-training and fine-tuning in Section \ref{results:model_quality}. It indicates that before a loss spike appears, SDCs may have already affected model training for an unknown period.

\subsection{Mitigating the Impact of SDCs}
Given its silent nature, it is important to mitigate the impact of SDC on LLM training. One direction is timely SDC detection. We examine if algorithm-based-fault tolerance (ABFT), specifically check-summed matrix multiplication \cite{utexas_abft}, could be used at the submodule level to detect SDCs, with the full experiment details in Appendix \ref{appendix:abft}.  We find that ABFT fails to reliably detect SDCs with high precision and recall: therefore, additional recomputation may be the best solution to reliably detect SDCs in practice. One possible approach is using an additional shadow data-parallel replica during training. At each training step, the shadow replica chooses a target data-parallel rank whose inputs are used for its computation. The gradients from the shadow data-parallel replica and the target data-parallel rank can be compared during gradient all-reduce over data-parallel ranks to identify if an SDC has occurred. This idea is aligned with the SDC scanners on hot standbys used in training Gemini \cite{geminiteam2024geminifamilyhighlycapable}. 

Another direction is to mitigate the impact of SDCs on the training trajectory and the model quality when SDCs are not detected timely. As discussed in Section \ref{results:model_quality}, the impact of SDCs on model quality is closely related to the sharpness of the loss surface. Therefore, one future direction is to conduct a deeper analysis of the connections between loss spikes and SDC. Future work could examine whether methods that reduce the sharpness of the loss surface or avoid optimization towards sharp regions \cite{lee2023explicitcurvatureregularizationdeep, bahri2022sharpnessawareminimizationimproveslanguage} can help reduce the divergence of model parameters and loss spikes caused by SDCs.

\section{Conclusion}
In this work, we propose a study setting that enables us to thoroughly investigate the effects of real-world SDC on LLM training. Pairing healthy and unhealthy nodes together using our synchronization mechanisms, we isolate and examine the impacts of SDCs at different levels. 
We show that although in most cases the perturbations from SDCs on submodule computation and gradients are relatively small, SDCs can make models converge to different optima with different weights. We further show that SDCs can be evasive if we only monitor the training loss while in the extreme case, they can cause training loss spikes and fully corrupt the model. Our study reveals that the impact of SDC varies on different unhealthy nodes and is closely related to the loss surface of the training task.
Our work further provides concrete insights for improving SDC detection and mitigating the impact of SDC in the future.

\section{Limitations}
We note a few limitations in this work. First, the number of unhealthy nodes we could access for our experiments was restricted because unhealthy nodes are rare due to rigorous manufacturer testing and filtering of hardware components before reaching the data-center.
The unhealthy nodes retained after detection by the production fleet management flow and used in our study \emph{were only temporarily held for the purpose of our study}. They are subsequently being repaired and returned to the production pool.

Second, our study focuses on training with tensor parallelism only on a single node. Since each node exhibits different degrees of SDC, we could not run large-scale training with more complex parallelism strategies for every unhealthy node as we did in the tensor parallel setting due to resources and limited time. However, our experimental setting is still meaningful for understanding SDCs in LLM training. For large-scale LLM training, tensor parallelism is commonly adopted with the degree set to be the number of accelerators in a single node \cite{narayanan2021efficient}, which makes our setting highly relevant. Furthermore, in our work, each model’s training is fully computed on an unhealthy node, which makes the effects of SDCs more visible.

Third, our work shows a necessary trade-off between fine-grained analysis and SDC occurrence. Our synchronization mechanisms inevitably introduce additional overhead and reduce accelerator utilization, which can lead to a decrease in the frequency of SDCs. As shown in Figure \ref{fig:attention_forward_mismatch_frequency_over_time} where we use computation synchronization at each submodule computation, Node 14 exhibits SDCs very infrequently over 4000 optimizer steps. By contrast, in Figure \ref{fig:fix_weight_only} where we use parameter synchronization at each step, the same node shows mismatched gradient norms at every step. The difference in SDC occurrence on the same node suggests more synchronization will decrease the accelerator utilization and further reduce the frequency of SDC. 
However, this trade-off is necessary to analyze SDCs at a finer granularity apart from high-level metrics like loss curves or gradient norms. Our synchronization mechanism allows us to characterize the impact of SDC at different granularities because we need to prevent the SDC error from being propagated to the next measurement. We will study how to mitigate the impact of this trade-off and propose more effective methods to analyze SDCs in the future.

Finally, we only observe the loss spikes in some fine-tuning experiments but not in our pre-training experiments. However, loss spikes can occur in practice during pre-training  \cite{chowdhery2023palm}. This discrepancy might be because the number of optimizer steps is not large enough to enter into a region where SDCs cause loss spikes or because the size of our model is not large enough. We also note that it can be challenging to reproduce and investigate the loss spikes caused by SDCs due to the randomness of SDCs. As observed in our work, if we monitor intermediate computation tensors during training, it decreases accelerator utilization, which affects the frequency of SDC and potentially prevents the occurrence of loss spikes. Future work can continue pre-training for a longer period on unhealthy nodes, potentially in a multi-node setting, to characterize the relation of loss spikes or NaN issues with SDC.

\section*{Acknowledgments}
We thank Thomas Fussell, Catalin Gabriel Manciu, Alexander Zhipa, Tushar Sharma, Manish Reddy, Stan Ivashkevich, Mohammad El-Shabani, Dave Goodell and Ron Diamant for providing advice.

\bibliography{custom}

\appendix

\section{Additional Background on SDCs}
\label{appendix:sdc_related_work}

We further discuss background and prior work on detecting and correcting silent data corruption (SDC). Generally, SDCs cannot be guarded against effectively by common mechanisms like ECC memory \cite{assetintertechSilentData} as they arise during computation. Therefore, several prior works have investigated SDC detection and correction.

\textbf{Simulating SDC Faults.} 
Some prior works focus on understanding how simulated SDC-like faults propagate through training and inference in deep learning systems. For example, \citet{heunderstandingandmitigatinghardwarefailures2023} simulate hardware faults during training of several different deep learning architectures, characterize the error modes that result from simulated faults and error propagating through model training, and understand the conditions in which these faults must occur to destabilize training. However, relative to modern LLMs, the model sizes in their experiments are small and the training workloads are different compared with LLMs. Although the simulation is comprehensive, it is unknown how real-world SDC would affect LLM training in practice. In contrast, our work uses the real-world unhealthy hardware to train Transformer models with billions of parameters to understand how real-world SDCs behave in LLM training.

\textbf{Detection via Aggregate Statistics.} Some prior works focus on SDC detection primarily through only monitoring aggregate training statistics. For example, \citet{hepermanenthardwarefailures} examines monitoring Inf/NaN results and loss spikes during training to identify when SDCs occur. Likewise, \citet{madrdna2024} detect SDCs with high precision using error signatures derived from analyzing the distribution of model neuron activations.

\textbf{Soft Redundancy and Protecting Inference Computation.} However, since SDCs can also occur silently without impacting aggregate quantities like loss or gradient norm, other works add minor levels of redundancy to detect SDCs with higher precision. For example, algorithm based fault tolerance (ABFT) approaches compute low overhead checksum computations alongside the original operation to check against \cite{wupanruosdcresilientabftmatmul, zhaiftblas}.
Most prior work using ABFT examines using detection in safety-critical applications \cite{Kosaian_2021} and specifically to protect deep learning inference computations\cite{Zhao_2021}. More recent work examined using this to protect the inference of vision transformers \cite{xue2023approxabftapproximatealgorithmbasedfault}.

\textbf{Detection with Exact Recomputation.} Finally, some works fully recompute values to check for SDC occurrence during training. For example, Gemini uses additional hardware to scan for SDCs and isolates incorrect computations by deterministically replaying computations \cite {geminiteam2024geminifamilyhighlycapable}.

\section{Model Pretraining Details} \label{appendix:model_training}
\subsection{Dataset and Preprocessing}


In this section, we describe the dataset preprocessing done during the training experiments described in Sections \ref{sec:primitive}, \ref{sec:sdc_single_optimizer_step}, \ref{sec:multiple_training_steps} and Appendix Section \ref{appendix:abft}. We train on the \verb|20220301.en| split of the Wikipedia dataset \cite{wikidump} with a sequence length of $L=4096$ tokens. Using the full \verb|20220301.en| split, we first tokenize the dataset using the pre-trained Byte-Pair Encoding (BPE) Llama-3 tokenizer \cite{dubey2024llama3herdmodels} . We then concatenate the entire token sequence and chunk the dataset into chunks of sequence length 4096 to maximize context and accelerator utilization during training. The entire epoch of sequences is then shuffled with a fixed random seed by sequence then saved to disk. At training time, using \verb|torch.distributed|, each tensor parallel rank ingests the same subset of data at the same time using a distributed dataloader via PyTorch/XLA's parallel dataloader\footnote{\url{https://github.com/pytorch/xla/blob/master/torch_xla/distributed/parallel_loader.py}}, which provides buffering to hide CPU to device data load latency.

\subsection{Model Architecture}

We use a Llama3-8B style model architecture \cite{dubey2024llama3herdmodels} trained from Kaiming and Xavier uniform initialization on the weights and biases, where each decoder layer has 32 self-attention heads with group query attention (GQA) over 8 Key-Value heads and an feed-forward network using SwiGLU. All models are trained using tensor parallelism and ZeRO-1 with sequence parallelism \cite{rajbhandari2020zeromemoryoptimizationstraining, korthikanti2022reducingactivationrecomputationlarge} with the XLA backend corresponding to the respective healthy-unhealthy node pair.

In this section, we describe the model architecture trained in the experiments described in Sections \ref{sec:primitive}, \ref{sec:sdc_single_optimizer_step}, \ref{sec:multiple_training_steps} and Appendix Section \ref{appendix:abft}. In the experiments described in Sections \ref{sec:primitive} and Appendix Section \ref{appendix:abft}, the model trained contains only 16 decoder block layers (and half the number of parameters as the Llama3-8B configuration), while in Sections \ref{sec:sdc_single_optimizer_step} and \ref{sec:multiple_training_steps}, the model trained contains 32 decoder block layers and is equivalent to the Llama3-8B model configuration (with number of parameters). The details of each decoder block are given below, with sequence length $n=4096$ and token dimension $d=4096$:
\begin{enumerate}
    \item Given an input of token embeddings $X \in \mathbb{R}_{n \times d}$ sharded by sequence length (dimension $n$), we perform an all-gather such that each TP rank has the entire input $X$ and perform Group Query Attention (GQA)  \cite{ainslie2023gqatraininggeneralizedmultiquery} with 32 heads, head dimension of $d / 32 = 128$, and 8 key-value heads.
    We do not use FlashAttention \cite{dao2022flashattentionfastmemoryefficientexact} for our model architecture, instead using a standard GQA implementation.
    \item After the concatenation of head results and subsequent output row-parallel linear projection, the results are reduce-scattered to sequence parallelism, such that each TP rank has an equal split of tokens. We add a residual connection (add the original token embeddings) followed by a Layer Normalization.
    \item We then enter the FFN primitive by all-gathering, such that each TP rank has the entire new input. We perform an standard FFN with Swish-Gated Linear Unit (SwiGLU) activation \cite{shazeer2020gluvariantsimprovetransformer}, projecting to an intermediate dimension of 16384 (4x), performing SwiGLU, then projecting back down into $d=4096$. We perform the FFN with gradient checkpointing \cite{chen2016trainingdeepnetssublinear} on the intermediate dimension to save memory and avoid needing to persist forward matrices of size $4096 \times 16384$ to HBM for the backwards pass.
\end{enumerate}

\subsection{Model Hyperparameters}

Our model training hyperparameters are given below. For the primitive investigation in Section \ref{sec:primitive} and Appendix Section \ref{appendix:abft}, we train at a global batch size of 16, due to increased throughput from cross-node communication, while for the single and multiple optimizer step settings in Sections \ref{sec:sdc_single_optimizer_step} and \ref{sec:multiple_training_steps}, we train at a global batch size of 256.
\begin{itemize}
    \item Sequence length: 4096
    \item Embedding dimension: 4096
    \item Sharding strategy: ZeRO-1 with sequence parallelism
    \item Optimizer: Adam with $\beta_1 = 0.9$ and $\beta_2$ = 0.99
    \item Weight decay ($L_2$ regularization): 0.01
    \item Learning rate (LR) schedule: linear aramup, cosine annealing
    \item Total Steps (for LR schedule): 100,000
    \item Warmup Steps (for LR schedule): 2,000
    \item Training precision: \verb|bfloat16|
    \item Mixed precision: False
    \item Micro-batch size (for gradient accum.): 1
    \item Gradient norm clipping with max norm 1.0
    \item Rounding mode: Round to Nearest
\end{itemize}

Specifically for Sections \ref{sec:primitive} (submodule investigation), we train $M=4500$ optimizer steps using mixed-precision Adam \cite{micikevicius2018mixedprecisiontraining} and a global batch size of $B=16$. Note that we use a smaller batch size and number of decoder layers than those in later experiments because computation synchronization brings additional cross-node communication and comparison, which greatly decreases training throughput and increases required memory usage.

Specifically for Sections \ref{sec:sdc_single_optimizer_step}, \ref{sec:multiple_training_steps} (gradient and synchronization-free setting), we train $M=2500$ optimizer steps using mixed-precision Adam \cite{micikevicius2018mixedprecisiontraining}.

\section{Submodule Investigation Implementation Details} \label{appendix:primivite_investigation}

\subsection{Submodule Investigation Integration with TorchXLA and Autograd}

To integrate the lock-step communication mesh used in the primitive investigation in Section \ref{sec:primitive} into the forwards and backwards computations of a LLM training run, we developed a set of `torch.autograd.Function` implementations, which implement the communication mesh as forwards or backwards behavior. We can then insert and call these functions at the locations in which we'd like to investigate the transformer primitive outputs. When \verb|ComparisonForwardAutograd| is called on a tensor, it does no checks in the backwards pass but checks and outputs the computes SDC infrequency and severity for that tensor in the forward pass. When \verb|ComparisonBackwardAutograd| is called on a tensor, it does no checks in the forwards pass but checks and outputs the computed SDC infrequency and severity for the gradient of the activation corresponding to that tensor (i.e. the tensor propogated backwards at that location in the backwards pass). For implementations, see Figure \ref{fig:auto_grad_implementation} below.

As in Figure \ref{fig:primitive_investigation_appendix}, we insert \verb|ComparisonFwdAutograd| calls at any of the red arrow locations, as we want to compare the computed forward tensor values for each corresponding set of TP ranks between healthy and unhealthy hosts prior to a reduce-scatter. Likewise, to check the values prior to the reduce scatter after backwards primitives, we insert calls to \verb|ComparisonBwdAutograd| at the locations of the blue arrows, so that autograd will communicate and compare the backward pass input-gradients and return mismatch statistics.

\begin{figure}[t]
    \centering
    \includegraphics[width=\linewidth]{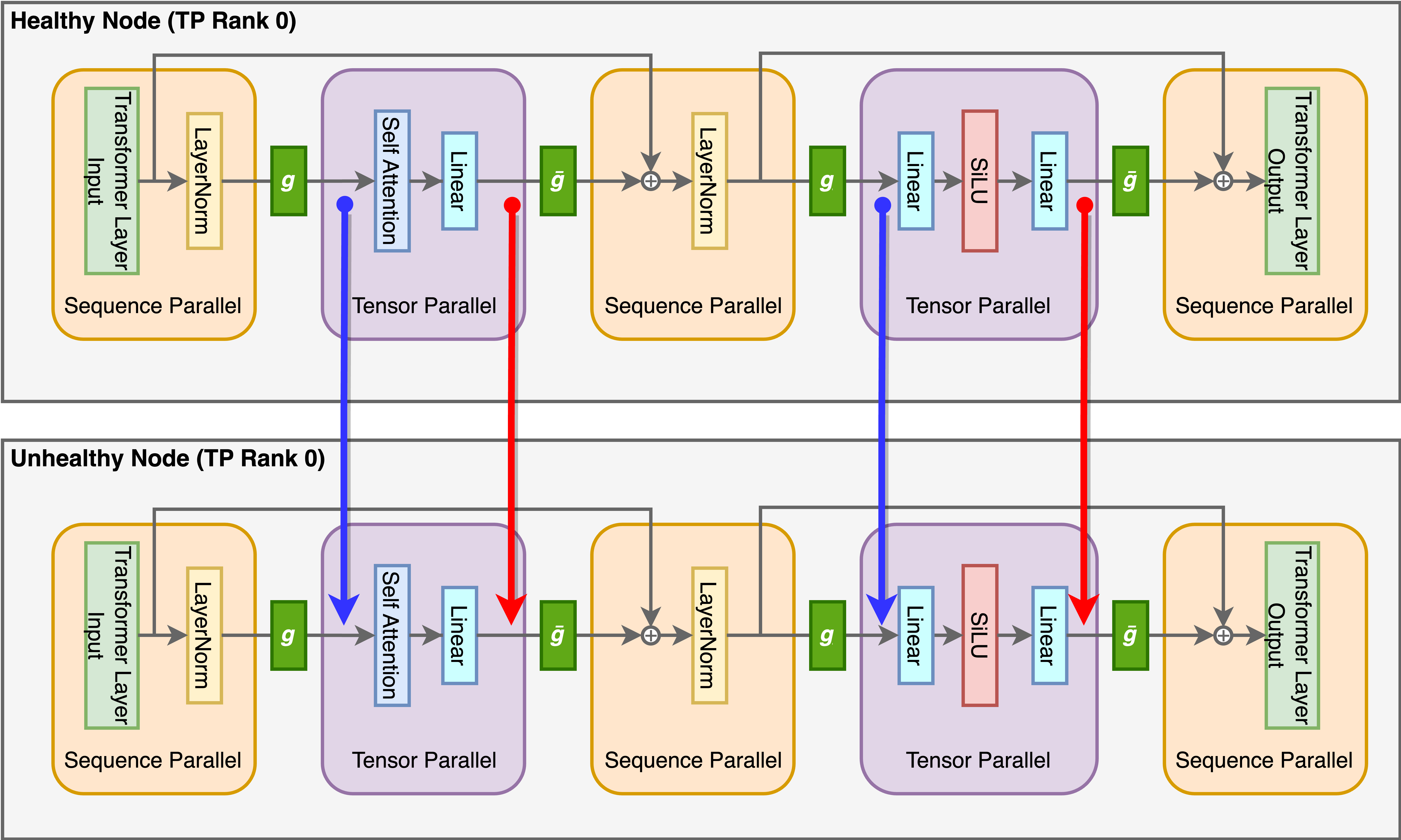}
    \vskip -0.05in
    \caption{Illustration of a transformer decoder block under our ``lockstep parallelism" in the primitive investigation setting, where the arrows indicate the intermediate tensors from the unhealthy corrected by corresponding tensors from the healthy host (red in forwards pass, blue in backwards pass) and where the PyTorch autograd functions in Figure \ref{fig:auto_grad_implementation} are inserted in implementation. Note that in the forward pass $g$ is an all-gather and $\bar{g}$ is a reduce-scatter, while in the backwards pass $g$ is an reduce-scatter and $\bar{g}$ is an all-gather.}
    \vskip -0.15in
    \label{fig:primitive_investigation_appendix}
\end{figure}

This implementation allows the PyTorch autograd engine to automatically handle our primitive investigation communication mesh as it computes forwards and backwards passes for the model.

\begin{figure*}[htbp]
    \begin{longfbox}[title=Autograd Implementation]{}
    \begin{small}
    \begin{verbatim}
@torch.no_grad()
def check_across_data_parallel_ranks(
  tensor_to_check: torch.tensor,
  check_args: CheckConfig,
  layer_count: int,
  tag: Optional[str] = None,
) -> Tuple[torch.Tensor, Optional[Dict[str, Any]]]:
  # For a given tensor:
  # (1) gathers the copy of the tensor on both the healthy and unhealthy host.
  # (2) computes the difference between them and calculates statistics on 
  #     infrequency/severity of mismatching values.
  # (3) returns the correctly computed version of the tensor (from the healthy 
  #     host) and computed error stats.
  ...
  return new_tensor, error_statistics

class ComparisonFwdAutograd(torch.autograd.Function):
  @staticmethod
  def forward(
    ctx, input_tensor, forward_args: CheckConfig, layer_count: int, tag: str
  ):
    # Check mismatching values and return error statistics in fwd pass.
    input_tensor, forward_error_statistics = check_across_data_parallel_ranks(
        input_tensor, forward_args, layer_count + 1, tag=tag
    )
    return input_tensor, forward_error_statistics

  @staticmethod
  def backward(ctx, grad_tensor, _unused_error_dict):
    # Do nothing on the bwd pass.
    return grad_tensor, None, None, None

class ComparisonBwdAutograd(torch.autograd.Function):
  @staticmethod
  def forward(
    ctx,
    input_tensor,
    backward_args: CheckConfig,
    layer_count: int,
    backwards_error_statistics: ErrorDict,
    prefix: str,
  ):
    # Do nothing in the fwd pass.
    ...
    return input_tensor

  @staticmethod
  def backward(ctx, grad_tensor):
    # Check mismatching values and return error statistics in bwd pass.
    new_grad_tensor, backwards_error_statistics = check_across_data_parallel_ranks(
        grad_tensor, ctx.backward_args, ctx.layer_count, tag=ctx.prefix
   )
   backwards_error_statistics = backwards_error_statistics.add_prefix(ctx.prefix)
   ctx.backwards_error_statistics.add_inplace(backwards_error_statistics)
   return new_grad_tensor, None, None, None, None
\end{verbatim}
    \end{small}
    \end{longfbox}
    \caption{Abbreviated implementation of helper autograd functions, which are inserted into transformer primitive locations prior to the reduce-scatter to analyze tensors of interest. The autograd engine then handles the primitive investigation communication as we compute forward and backwards passes.}
    \label{fig:auto_grad_implementation}
\end{figure*}

\subsection{Detailed Primitive Investigation Results} \label{appendix:primitive_investigation_detailed_results}

We generally do not observe any per decoder layer trends in our results and provide detailed breakdowns of mismatch frequency and severity below.

\subsubsection{Frequency of Mismatching Elements}
Detailed per node and per decoder results on the frequency of mismatching tensor elements in the forward and backwards passes are shown in Tables \ref{fig:forward_primitive_frequency} and \ref{fig:backwards_primitive_frequency}, respectively.

\subsubsection{Severity of Mismatching Elements}
Detailed per node results on the average maximum severity of mismatching tensor elements in the forward and backwards passes are shown in Tables \ref{fig:forward_primitive_max_severity} and \ref{fig:backwards_primitive_max_severity}, respectively.

\section{Algorithm Based Fault Tolerance (ABFT) Discussion}  \label{appendix:abft}

\subsection{Experiment Details}

\textbf{Experiment.} \emph{Can algorithm-based fault tolerance (ABFT) detect the errors in Transformer submodule outputs?}
Algorithm-based fault tolerance (ABFT) is one common scheme to detect SDC errors during deep learning computations \cite{Zhao_2021, xue2023approxabftapproximatealgorithmbasedfault}. Specifically, \citet{utexas_abft} propose adding a checksum to matrix multiplication to flag if SDCs have occurred during computation, comparing the results of two data paths, the matrix multiplication result and the checksum element. For a floating-point matrices $A \in \mathbb{R}^{m \times k}, B \in \mathbb{R}^{k\times n}$, $C = AB$ and a column vector of ones $w$, ABFT reports an SDC if:
\begin{equation}
    \label{equation:abft_condition}
    \| C w - A( Bw) \|_\infty > \tau  \| A \|_\infty \| B \|_\infty
\end{equation}
where $\tau = ku$ and $u$ is the unit-roundoff determined by the FP precision used in the matrix multiplication. 

In this experiment, we added ABFT into the computation of all linear layers in the forward and backwards pass of model training. For each kind of matrix multiplication, we record the frequency of ABFT-flags across TP ranks and layers using the condition in \ref{equation:abft_condition}. We train the Transformer model with $D=8$ decoder layers under the precision of \verb|float32|, with rest of hyperparameters the same as detailed in Appendix \ref{appendix:model_training}. Note that underlying theoretical assumptions mean that this ABFT method cannot be applied in lower precision data types like \verb|bf16|. More discussion and details can be found in below in Appendix Section \ref{appendix:abft_bf16}. 

\subsection{Experiment Results}

\begin{figure}[t]
    \vskip -0.1in
    \centering
    \includegraphics[width=0.85\linewidth]{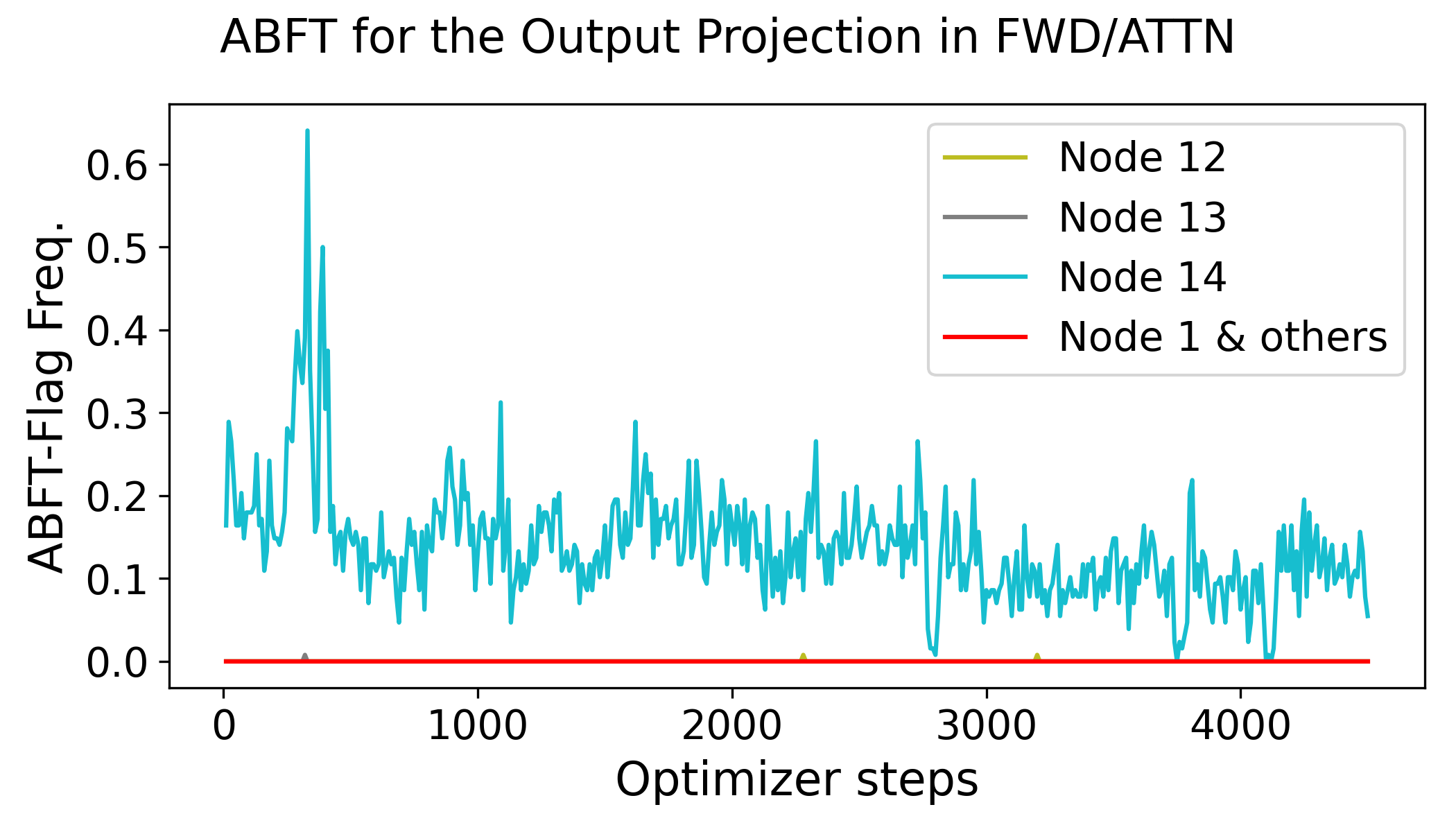}
    \vskip -0.1in
    \caption{Frequency of ABFT flaging an SDC in the forward computation of the output projection in the attention module during training.}
    \label{fig:abft_example}
    \vskip -0.1in
\end{figure}

We find ABFT fails to flag any error on most of the unhealthy nodes, except Node 14. In Figure \ref{fig:abft_example}, we observe ABFT only frequently flags SDCs on Node 14 while flagging little to no errors on other unhealthy nodes. We also note that the rates of SDCs flagged in this \verb|float32| ABFT setting do not necessarily correspond to submodule output mismatches at \verb|bfloat16| induced by SDCs in the Section \ref{sec:submodule_outputs_results} results. As from Figure \ref{fig:attention_forward_mismatch_frequency_over_time}, SDCs on Node 14 occur as a singular spike of mismatches at \verb|bfloat16| as observed in Section \ref{sec:submodule_outputs_results}, which conflicts with how often ABFT flags SDCs on the same node at \verb|float32|.

\subsection{Possible Explainations for ABFT Detection Failure} \label{appendix:submodule_vs_abft}

There are several possible reasons for why ABFT cannot reliably detect SDCs in our experiments. First, due to the low mismatch frequency, the overall impact on matrix norm could be smaller than floating point error bounds, which breaks the assumption of ABFT. Second, introducing ABFT changes the executed workload and decreases compute utilization in our implementation, which might also decrease the rate of SDC occurrence. Finally, SDCs may occur outside the ABFT-protected matrix multiplication. More detail on each is provided below:
\begin{itemize}
    \item \emph{Low frequency and severity of SDC-induced tensor mismatches:} As noted in the submodule results in Section \ref{sec:submodule_outputs_results}, the mismatch severity and frequency of silent data corruption errors results in low impact on matrix norms, which ABFT relies on to avoid flagging false positives. Assuming the SDC arises during matrix multiplication, the observed SDC occurrence is well within floating point error bounds and thus not flagged by ABFT methods. For example, for Node 10, we observe that 4.78$e$-3 of the forward attention output elements are perturbed by an worst-case factor of 1120 times. Using Eq. \ref{equation:abft_condition}, we see that, in the worst case, when all of these errors lie on a single row, one of the row-sum values in $Cw$ changes by an estimated factor of $4.78$e$-3 \times 1120 = 5.35$. Likewise, the LHS quantity $\|Cw - A(Bw)\|_\infty$ would be impacted by roughly the same estimated factor, which is less than an order of magnitude increase from the previous value and unlikely to cause a failure against the RHS threshold in Eq. \ref{equation:abft_condition}.
    \item \emph{ABFT overhead decreases accelerator utilization}: We observe that frequency of SDC occurrence is dependent on accelerator utilization adding ABFT overhead decreases utilization, which might decrease the rate of SDC occurrence. In Section \ref{sec:submodule_outputs_results}, our results on the frequency of SDC-induced mismatches show that SDC occurrence potentially is a function of system-level metrics like power draw and overall system-level utilization. ABFT decreases accelerator utilization and potentially changes this system-level profile, reducing SDC occurrence.
    \item \emph{SDCs occurring outside the matrix multiplication:} We hypothesized that checksummed matrix multiplication would be able to flag SDC errors due to matrix multiplications generally being the most compute-intensive and  highest utilization stage of a deep learning model. However, ABFT flagging no SDCs on known unhealthy hosts may suggest that, preconditioned on existing manufacturer testing and vetting, SDCs might more commonly arise outside of matrix multiplication.
\end{itemize}

\subsection{ABFT and Reduced Precision Datatypes} \label{appendix:abft_bf16}

We run our ABFT detection at \verb|float32| to respect the assumptions to derive the threshold limits for ABFT. Despite lower precision datatypes like \verb|bf16| and even \verb|fp8| are commonly used in LLM training \cite{lee2024fp8againquantifyingeffects}, the error bounds for ABFT are derived from IEEE-754 floating point (\verb|fp32| or higher precision) error bounds for matrix multiplication \cite{matrixcopmutations2013}. These bounds are derived under strict assumptions on the number of mantissa bits (and size of unit-roundoff), which do not hold for \verb|fp16| or \verb|bf16| datatypes. In the ABFT experiment, we ran our ABFT detection at \verb|float32|: when running the check as-is at \verb|bf16|, we observed several false positives on healthy nodes, confirming this conjecture. 

We elaborate on reasons why algorithm-based fault tolerance (ABFT), specifically checksummed matrix multiplication, required additional consideration to be deployed in real-world LLM training settings. Specifically, we noted that checksummed matrix multiplication of floating point matricies was based in bound on floating point error, which make several assumptions on the unit-roundoff or machine epsilon of computation data types.

Recall that \citet{panruowu_online_soft_error_correction} and \citet{utexas_abft} proposed adding a checksum row and column to two-sided matrix multiplication for online floating-point aware detection of silent data corruption errors in matrix multiplication. For some matrix product $C = A \times B$, column vector $w$, we say an checksum error (and SDC) has occurred if:
\begin{equation} \label{equation:abft_condition2}
    \| C w - A( Bw) \| > \tau  \| A \|_\infty \| B \|_\infty
\end{equation}
where $\tau = ku$, where $k$ is the contraction dimension of the matrix multiplication and $u$ is the unit-roundoff of the datatype precision used in the matrix multiplication. We can determine $u$ in PyTorch by using the function \verb|torch.finfo(<dtype>).eps|. The above threshold is derived under the assumptions of Equation 2.7.11 from \citet{matrixcopmutations2013}, a bound for round-off error in dot products, restated below for convenience.

\emph{Equation 2.7.11}: If $n\mathbf{u} \leq 0.01$ where $n$ is the size of the dot-product shared dimension, $\mathbf{u}$ is the unit round-off of the datatype used for computation, and $fl(x^T y)$ then
\begin{gather*}
    |fl(x^Ty) - x^t y| \leq 1.01 n \mathbf{u} |x|^T|y|
\end{gather*}
For types like \verb|float32| or \verb|float64| (and their corresponding \textbf{u} values) the assumption $n\mathbf{u} \leq 0.01$ is very reasonable. In our model configuration case, for a model with embedding dimension $4096$:
\begin{enumerate}
    \item \verb|float32|: $\mathbf{u}= 1.19215e-07, n\mathbf{u} \approx 0.0005 < 0.01$
    \item \verb|float16|: $\mathbf{u}= 0.0009765625, n\mathbf{u} \approx 4 \nless 0.01$
    \item \verb|bfloat16|: $\mathbf{u}= 0.0078125, n\mathbf{u} \approx 32 \nless 0.01$
\end{enumerate}
We see that for types less precise than \verb|float16|, ABFT error bound in Equation \ref{equation:abft_condition2} no longer always holds true and checksummed matrix multiplication methods would possibly raise false positives. We observed this empirically to be true, where the above threshold at \verb|bfloat16| raised false positive flags on healthy nodes that passed production margin tests, while raising no errors on the same node at \verb|float32|. Thus, more work in error analysis is needed to extend ABFT methods properly to low precision datatypes.

\section{Finetuning Experiment Details} \label{appendix:finetuning}

\subsection{Dataset Details}
Details and a brief description of each dataset are provided below:
\begin{enumerate}
    \item ScienceQA \cite{lu2022learn} is a multiple-choice question answering (QA) dataset of grade school science and humanities questions: note that we remove all questions with an image context.
    \item BoolQ \cite{clark2019boolqexploringsurprisingdifficulty} is a QA dataset for naturally occuring yes/no questions each with a page of contextual and relevant information.
    \item OpenbookQA \cite{OpenBookQA2018} is a dataset containing multiple-choice questions each with context modeled like an open-book exam, requiring multi-step reasoning, use of additional common and commonsense knowledge, and rich text comprehension. 
    \item MathQA \cite{mathqadataset} is a datset of English multiple-choice math word problems covering multiple math domain categories.
    \item RACE \cite{raceLai2017large} is a large-scale reading comprehension multiple-choice question answering dataset, collected from English examinations in China, which are designed for middle school and high school students. 
\end{enumerate}

\subsection{Finetuning Prompts}
In Figure \ref{fig:finetuning_prompt_structure}, we show our prompts for finetuning, which are structrued in the following general form. During training we include the full context including the correct answer, while during evaluation we remove everything after \verb|### CORRECT ANSWER| and ask the model to continue generating the answer.
\begin{figure}[htbp]
    \begin{longfbox}[title=Finetuning Prompt Structure]{}
    \begin{small}
    \begin{verbatim}
### QUESTION
{question}

### CONTEXT
{context}

### CHOICES
A: {choice1}
B: {choice2}
...

### CORRECT ANSWER
A: {choice 1}
\end{verbatim}
    \end{small}
    \end{longfbox}
    \caption{General finetuning prompt structure. }
    \label{fig:finetuning_prompt_structure}
\end{figure}

\subsection{Masking and Padding Details}

For finetuning, we right-pad sequences to a fixed sequence length of 2048 tokens (with the exception of 4096 for the RACE dataset) using the Mistral tokenizer \verb|EOS| token. Furthermore, we mask out padding tokens during training so that they do not contribute to loss and gradient computation by setting the \verb|labels| for padding token positions to \verb|-100|, so that they are ignored by PyTorch cross-entropy loss calculation \footnote{\url{https://pytorch.org/docs/stable/generated/torch.nn.CrossEntropyLoss.html}}.
\subsection{Hyperparameters and Finetuning Configuration}

We use the HuggingFace \verb|optimum| package to finetune the \verb|mistralai/Mistral-7B-v0.3| model on our experiment nodes using tensor parallelism and ZeRO-1 optimizer with sequence parallelism. For finetuning, we use the following hyperparameters.

\begin{itemize}
    \item Sequence length: 4096 for RACE, 2048 otherwise
    \item Sharding strategy: ZeRO-1 with sequence parallelism
    \item Optimizer: Adam with $\beta_1 = 0.9$ and $\beta_2 = 0.999$
    \item Weight decay ($L_2$ regularization): 0.001
    \item Learning rate schedule: constant with linear warmup.
    \item Warmup Steps (for computing cosine LR scheduler): 5\% of total steps
    \item Training precision: \verb|bfloat16|
    \item Global batch size: 32
    \item Micro-batch size (for gradient accum.): 1
    \item Gradient norm clipping with max norm 0.3
    \item Rounding mode: Round to Nearest
\end{itemize}

For learning rates, we chose the following values, noted alongside their training dataset size. These were tuned such that the finetuning on the train split improves the model on evaluation set performance on a healthy, non-SDC producing node.
\begin{table}[h]

\caption{Finetuning learning rate (LR) and dataset size for each dataset used in Experiment \ref{experiment:finetuning}.}
\vskip -0.15in

\begin{center}
\begin{small}
\begin{sc}
\setlength{\tabcolsep}{3pt}
\renewcommand{\arraystretch}{1.2} 
\begin{tabular}{ccc}
\hline
\toprule
Dataset & LR & Dataset Size (\# train seqs.) \\
\midrule
BoolQ & 5$e$-6 & 9,430 \\
CosmosQA & 5$e$-6 & 25,260 \\
MathQA & 1$e$-5 & 29,837 \\
OpenbookQA & 2$e$-6 & 11,920 \\
ScienceQA & 1$e$-6 & 12,700 \\
RACE & 1$e$-5 & 87,900 \\
\bottomrule
\end{tabular}

\end{sc}
\end{small}
\end{center}

\vskip -0.2in
\label{fig:finetuning_lr_data_size}

\end{table}

\subsection{Full Finetuning Test Accuracy and Disagreement Percentage Results} \label{appendix:full_finetuning_results}

The full table of finetuning results across all datasets is shown below in Table \ref{tab:finetuning_results}. We again note that across all datasets, finetuned models on unhealthy nodes achieve improved performance over the unfinetuned baseline and similar (though differing performance) compared to models deterministically finetuned on healthy nodes. Specficially, we observe, that this disagreement percentage (or delta from the healthy node finetuning) is comparable to finetuning under a differnt dataset shuffling seed.

\begin{table*}[t]

\begin{center}
\begin{tiny}
\begin{sc}

\begin{tabular}{ccccccccccccc}
\toprule
\multirow{2}{*}{Configuration} & \multicolumn{2}{c}{CosmosQA} & \multicolumn{2}{c}{MathQA} & \multicolumn{2}{c}{ScienceQA} & \multicolumn{2}{c}{OpenbookQA} & \multicolumn{2}{c}{BoolQ} & \multicolumn{2}{c}{RACE} \\
 & TA & DP & TA & DP & TA & DP & TA & DP & TA & DP & TA & DP\\
\midrule
without fine-tuning & 56.33 & 44.80 & 24.02 & 82.35 & 72.62 & 34.22 & 74.70 & 29.30 & 70.21 & 27.89 & 73.63 & 25.44 \\
\midrule
Healthy Node & 90.79 & - & 37.22 & - & 84.17 & - & 83.80 & - & 90.06 & - & 87.47 & - \\
Healthy Node (Seed=$43$) & 89.50 & 6.70 & 38.83 & 56.75 & 87.90 & 14.57 & 86.30 & 18.70 & 90.31 & 5.26 & 87.62 & 9.95 \\
\midrule
Unhealthy Node 1 & 90.53 & 5.15 & 36.78 & 42.24 & 86.74 & 13.26 & 85.00 & 16.80 & 89.94 & 3.49 & 87.82 & 6.61  \\
Unhealthy Node 2 & 90.79 & 0.00 & 37.22 & 0.00 & 84.17 & 0.00 & 83.80 & 0.00 & 90.06 & 0.00 & 87.47 & 0.00 \\
Unhealthy Node 3 & 90.77 & 4.96 & 34.47 & 41.84 & 85.79 & 14.61 & 83.40 & 16.10 & 90.21 & 3.64 & 87.47 & 0.00 \\
Unhealthy Node 4 & 90.32 & 5.59 & 36.42 & 37.76 & 85.48 & 15.38 & 84.30 & 16.60 & 90.61 & 4.10 & 88.12 & 6.75  \\
Unhealthy Node 5 & 90.79 & 0.00 & 37.19 & 34.91 & 84.17 & 0.00 & 85.10 & 14.10 & 90.06 & 0.00 & 87.47 & 6.45 \\
Unhealthy Node 6 & 0.00 & 100.00 & 36.92 & 36.82 & 84.26 & 2.52 & 84.70 & 16.30 & 90.03 & 2.66 & 0.00 & 100.00 \\
Unhealthy Node 7 & 90.32 & 4.99 & 37.22 & 0.00 & 85.21 & 17.81 & 83.80 & 0.00 & 90.80 & 3.36 & 87.41 & 6.36  \\
Unhealthy Node 8 & 89.84 & 6.23 & 38.22 & 36.62 & 85.66 & 12.14 & 85.20 & 15.50 & 90.49 & 3.00 & 87.11 & 6.61 \\
Unhealthy Node 9 & 89.97 & 5.38 & 35.78 & 37.49 & 85.30 & 21.49 & 85.00 & 17.40 & 90.12 & 3.43  & 87.41 & 6.57 \\
Unhealthy Node 10 & 89.97 & 4.93 & 37.05 & 37.05 & 88.31 & 15.15 & 84.10 & 17.20 & 90.64 & 3.09  & 87.94 & 6.59 \\
Unhealthy Node 11 & 90.61 & 4.54 & 36.82 & 38.53 & 85.57 & 16.41 & 87.10 & 15.60 & 90.31 & 3.73  & 87.15 & 6.95 \\
Unhealthy Node 12 & 90.79 & 0.00 & 37.22 & 0.00 & 84.17 & 0.00 & 83.80 & 0.00 & 90.06 & 0.00  & 87.47 & 0.00\\
Unhealthy Node 13 & 90.79 & 0.00 & 37.22 & 0.00 & 84.17 & 0.00 & 83.80 & 0.00 & 90.06 & 0.00  & 87.11 & 6.77 \\
Unhealthy Node 14 & 89.74 & 6.12 & 38.63 & 32.29 & 84.17 & 0.00 & 85.00 & 15.90 & 90.06 & 0.00  & 87.62 & 6.40 \\
Unhealthy Node 15 & 90.77 & 3.27 & 38.53 & 40.87 & 84.17 & 0.00 & 83.80 & 0.00 & 90.06 & 0.00  & 87.39 & 6.61 \\
\bottomrule
\end{tabular}

\end{sc}
\end{tiny}
\end{center}

\caption{Finetuning results for six question answering tasks on different nodes. For each task, we report the test accuracy (TA) and the disagreement percentage (DP).  By default, the random seed for data shuffling is set as $42$. }
\label{tab:finetuning_results}

\vskip -0.15in
\end{table*}

\begin{table*}[t]

\begin{center}
\begin{tiny}
\begin{sc}

\begin{tabular}{rrrrrrrrrrrr}
\toprule
& & \multicolumn{9}{c}{SDC Node ID} \\
\midrule
Decoder Layer & Primitive  & Node 1       & Node 4         & Node 6        & Node 7       & Node 8        & Node 9       & Node 10      & Node 11      & Node 14       \\
\midrule
\multirow{2}{*}{1}     & fwd/attn         & 2.84$e$-5 & 3.31$e$-9  & 0.00$e$+0  & 1.02$e$-6 & 1.59$e$-9  & 2.92$e$-6 & 0.00$e$+0 & 1.43$e$-2 & 0.00$e$+0  \\
        & fwd/ffn          & 5.17$e$-7 & 4.37$e$-11 & 0.00$e$+0  & 5.18$e$-7 & 9.95$e$-12 & 1.52$e$-7 & 9.75$e$-6 & 2.40$e$-3 & 0.00$e$+0  \\
\midrule
\multirow{2}{*}{2}       & fwd/attn         & 1.45$e$-5 & 2.47$e$-9  & 3.89$e$-9  & 2.94$e$-6 & 1.70$e$-9  & 9.83$e$-6 & 8.62$e$-4 & 3.89$e$-2 & 1.82$e$-10 \\
        & fwd/ffn          & 2.86$e$-7 & 1.02$e$-10 & 2.39$e$-14 & 3.48$e$-8 & 1.42$e$-11 & 2.58$e$-7 & 2.18$e$-4 & 2.04$e$-3 & 0.00$e$+0  \\
\midrule
\multirow{2}{*}{3}      & fwd/attn         & 9.94$e$-6 & 1.78$e$-9  & 0.00$e$+0  & 1.76$e$-6 & 4.44$e$-9  & 6.31$e$-6 & 1.78$e$-3 & 2.67$e$-2 & 0.00$e$+0  \\
        & fwd/ffn          & 5.21$e$-7 & 1.42$e$-10 & 0.00$e$+0  & 9.43$e$-8 & 6.45$e$-12 & 3.74$e$-7 & 4.75$e$-4 & 2.24$e$-3 & 0.00$e$+0  \\
\midrule
\multirow{2}{*}{4}     & fwd/attn         & 8.82$e$-6 & 2.40$e$-9  & 0.00$e$+0  & 2.02$e$-6 & 3.86$e$-9  & 5.24$e$-6 & 1.63$e$-3 & 2.02$e$-2 & 2.61$e$-10 \\
        & fwd/ffn          & 4.83$e$-7 & 1.05$e$-10 & 0.00$e$+0  & 5.85$e$-8 & 1.85$e$-11 & 3.84$e$-7 & 4.66$e$-4 & 2.38$e$-3 & 0.00$e$+0  \\
\midrule
\multirow{2}{*}{5}      & fwd/attn         & 1.10$e$-5 & 2.99$e$-9  & 0.00$e$+0  & 1.99$e$-6 & 2.45$e$-9  & 6.85$e$-6 & 2.77$e$-3 & 2.34$e$-2 & 3.11$e$-10 \\
        & fwd/ffn          & 4.96$e$-7 & 9.04$e$-11 & 0.00$e$+0  & 1.07$e$-7 & 6.52$e$-12 & 3.53$e$-7 & 6.79$e$-4 & 2.19$e$-3 & 0.00$e$+0  \\
\midrule
\multirow{2}{*}{6}       & fwd/attn         & 1.23$e$-5 & 3.57$e$-9  & 0.00$e$+0  & 2.13$e$-6 & 5.14$e$-10 & 8.70$e$-6 & 3.98$e$-3 & 2.59$e$-2 & 0.00$e$+0  \\
        & fwd/ffn          & 6.07$e$-7 & 1.27$e$-10 & 8.67$e$-11 & 1.35$e$-7 & 2.77$e$-11 & 4.48$e$-7 & 9.17$e$-4 & 2.36$e$-3 & 0.00$e$+0  \\
\midrule
 \multirow{2}{*}{7}        & fwd/attn         & 1.05$e$-5 & 1.29$e$-9  & 2.33$e$-8  & 1.78$e$-6 & 1.03$e$-8  & 5.37$e$-6 & 4.39$e$-3 & 2.15$e$-2 & 0.00$e$+0  \\
        & fwd/ffn          & 6.27$e$-7 & 9.07$e$-11 & 0.00$e$+0  & 1.20$e$-7 & 7.15$e$-11 & 4.72$e$-7 & 1.18$e$-3 & 2.36$e$-3 & 0.00$e$+0  \\
\hline
\multirow{2}{*}{8}        & fwd/attn         & 1.69$e$-5 & 3.78$e$-9  & 0.00$e$+0  & 2.14$e$-6 & 4.47$e$-9  & 1.13$e$-5 & 5.08$e$-3 & 2.91$e$-2 & 0.00$e$+0  \\
        & fwd/ffn          & 5.01$e$-7 & 1.28$e$-10 & 0.00$e$+0  & 1.43$e$-7 & 1.24$e$-11 & 5.12$e$-7 & 1.19$e$-3 & 2.32$e$-3 & 0.00$e$+0  \\
\midrule
\multirow{2}{*}{9}       & fwd/attn         & 1.61$e$-5 & 1.04$e$-8  & 2.39$e$-14 & 2.45$e$-6 & 2.12$e$-9  & 1.70$e$-5 & 6.94$e$-3 & 3.63$e$-2 & 1.02$e$-10 \\
        & fwd/ffn          & 6.52$e$-7 & 1.03$e$-10 & 0.00$e$+0  & 1.27$e$-7 & 6.00$e$-11 & 5.22$e$-7 & 1.44$e$-3 & 2.23$e$-3 & 0.00$e$+0  \\
\midrule
\multirow{2}{*}{10}        & fwd/attn         & 1.36$e$-5 & 1.74$e$-9  & 8.56$e$-11 & 2.34$e$-6 & 8.23$e$-9  & 1.45$e$-5 & 6.57$e$-3 & 3.20$e$-2 & 0.00$e$+0  \\
        & fwd/ffn          & 5.62$e$-7 & 6.93$e$-11 & 0.00$e$+0  & 1.38$e$-7 & 8.89$e$-12 & 5.67$e$-7 & 1.40$e$-3 & 2.19$e$-3 & 0.00$e$+0  \\
\midrule
\multirow{2}{*}{11}      & fwd/attn         & 1.96$e$-5 & 4.25$e$-9  & 0.00$e$+0  & 2.29$e$-6 & 7.92$e$-11 & 1.46$e$-5 & 6.78$e$-3 & 3.45$e$-2 & 0.00$e$+0  \\
        & fwd/ffn          & 5.03$e$-7 & 7.97$e$-11 & 8.68$e$-11 & 9.69$e$-8 & 8.55$e$-12 & 6.67$e$-7 & 1.36$e$-3 & 2.23$e$-3 & 0.00$e$+0  \\
\midrule
\multirow{2}{*}{12}      & fwd/attn         & 1.97$e$-5 & 3.46$e$-9  & 0.00$e$+0  & 2.58$e$-6 & 2.03$e$-9  & 1.50$e$-5 & 6.81$e$-3 & 3.49$e$-2 & 0.00$e$+0  \\
        & fwd/ffn          & 5.63$e$-7 & 7.24$e$-11 & 8.88$e$-11 & 1.10$e$-7 & 9.51$e$-12 & 6.78$e$-7 & 1.35$e$-3 & 2.38$e$-3 & 0.00$e$+0  \\
\midrule
\multirow{2}{*}{13}      & fwd/attn         & 2.28$e$-5 & 9.72$e$-9  & 8.21$e$-11 & 3.01$e$-6 & 4.83$e$-14 & 1.86$e$-5 & 7.05$e$-3 & 4.09$e$-2 & 8.56$e$-11 \\
        & fwd/ffn          & 4.35$e$-7 & 8.06$e$-11 & 0.00$e$+0  & 5.00$e$-8 & 5.72$e$-12 & 6.67$e$-7 & 1.26$e$-3 & 2.20$e$-3 & 0.00$e$+0  \\
\midrule
\multirow{2}{*}{14}      & fwd/attn         & 1.78$e$-5 & 3.91$e$-9  & 4.79$e$-14 & 2.37$e$-6 & 4.67$e$-9  & 1.47$e$-5 & 7.66$e$-3 & 3.50$e$-2 & 0.00$e$+0  \\
        & fwd/ffn          & 3.83$e$-7 & 6.99$e$-11 & 2.39$e$-14 & 2.84$e$-8 & 8.84$e$-12 & 7.20$e$-7 & 1.49$e$-3 & 2.13$e$-3 & 0.00$e$+0  \\
\midrule
\multirow{2}{*}{15}       & fwd/attn         & 1.39$e$-5 & 1.33$e$-9  & 0.00$e$+0  & 1.75$e$-6 & 2.81$e$-9  & 1.31$e$-5 & 7.18$e$-3 & 2.61$e$-2 & 0.00$e$+0  \\
        & fwd/ffn          & 4.66$e$-7 & 8.00$e$-11 & 0.00$e$+0  & 8.01$e$-8 & 4.49$e$-11 & 6.76$e$-7 & 1.65$e$-3 & 2.20$e$-3 & 0.00$e$+0  \\
\midrule
\multirow{2}{*}{16}      & fwd/attn         & 1.25$e$-5 & 4.28$e$-9  & 4.79$e$-14 & 1.43$e$-6 & 2.02$e$-9  & 1.11$e$-5 & 6.93$e$-3 & 2.43$e$-2 & 9.60$e$-11 \\
        & fwd/ffn          & 4.86$e$-7 & 8.85$e$-11 & 0.00$e$+0  & 4.40$e$-8 & 5.02$e$-12 & 6.33$e$-7 & 1.46$e$-3 & 2.11$e$-3 & 0.00$e$+0 \\
\bottomrule
\end{tabular}

\end{sc}
\end{tiny}
\end{center}
\vskip 0.15in

\caption{Results for mismatch frequency for each transformer primitive forward, separated by decoder layer and averaged across all microsteps. The unhealthy Nodes 2, 3, 5, 12, 13 and 15 did not exhibit any mismatching tensors in forward passes in this experimental setting and thus are excluded from the table.}
\label{fig:forward_primitive_frequency}
\end{table*}
\begin{table*}[t]

\begin{center}
\begin{tiny}
\begin{sc}

\setlength{\tabcolsep}{3pt}
\begin{tabular}{rrrrrrrrrrrrrr}
\toprule
& & \multicolumn{12}{c}{SDC Node ID} \\
\midrule
Decoder & Primitive & Node 1        & Node 4        & Node 5        & Node 6        & Node 7        & Node 8        & Node 9        & Node 10       & Node 11       & Node 13       & Node 14       & Node 15       \\
\midrule
\multirow{2}{*}{16}           & bwd/ffn   & 3.12$e$-06 & 5.89$e$-11 & 6.21$e$-13 & 0 & 7.04$e$-08 & 2.55$e$-09 & 3.71$e$-08 & 7.89$e$-05 & 9.92$e$-05 & 4.91$e$-11 & 1.44$e$-09 & 7.09$e$-14 \\
              & bwd/attn  & 2.11$e$-04 & 3.08$e$-09 & 0 & 0 & 4.44$e$-06 & 1.40$e$-07 & 5.57$e$-06 & 2.59$e$-03 & 7.19$e$-03 & 0 & 3.09$e$-10 & 0 \\
\midrule
\multirow{2}{*}{15}             & bwd/ffn   & 3.11$e$-06 & 6.19$e$-11 & 1.58$e$-12 & 0 & 7.21$e$-08 & 2.46$e$-09 & 3.87$e$-08 & 8.45$e$-05 & 1.06$e$-04 & 9.57$e$-11 & 7.33$e$-10 & 0 \\
              & bwd/attn  & 1.81$e$-04 & 2.10$e$-09 & 0 & 0 & 4.07$e$-06 & 1.22$e$-07 & 5.07$e$-06 & 2.43$e$-03 & 6.56$e$-03 & 0 & 0 & 0 \\
\midrule
\multirow{2}{*}{14}             & bwd/ffn   & 3.09$e$-06 & 1.87$e$-11 & 1.46$e$-12 & 2.13$e$-11 & 6.89$e$-08 & 2.43$e$-09 & 4.12$e$-08 & 8.10$e$-05 & 1.02$e$-04 & 1.33$e$-10 & 2.14$e$-09 & 0 \\
              & bwd/attn  & 2.63$e$-04 & 2.24$e$-09 & 0 & 0 & 5.96$e$-06 & 1.62$e$-07 & 6.75$e$-06 & 2.78$e$-03 & 8.51$e$-03 & 0 & 1.64$e$-11 & 0 \\
\midrule
\multirow{2}{*}{13}             & bwd/ffn   & 3.09$e$-06 & 2.08$e$-11 & 4.06$e$-12 & 4.74$e$-10 & 7.04$e$-08 & 2.45$e$-09 & 4.02$e$-08 & 8.03$e$-05 & 1.01$e$-04 & 3.24$e$-10 & 9.38$e$-10 & 2.36$e$-14 \\
              & bwd/attn  & 2.85$e$-04 & 8.18$e$-09 & 0 & 6.14$e$-10 & 6.88$e$-06 & 1.89$e$-07 & 7.73$e$-06 & 2.70$e$-03 & 8.91$e$-03 & 0 & 4.79$e$-14 & 0 \\
\midrule
\multirow{2}{*}{12}            & bwd/ffn   & 3.08$e$-06 & 3.69$e$-11 & 1.62$e$-12 & 3.45$e$-11 & 7.37$e$-08 & 2.43$e$-09 & 4.19$e$-08 & 8.54$e$-05 & 1.13$e$-04 & 1.23$e$-10 & 6.67$e$-10 & 0 \\
              & bwd/attn  & 2.48$e$-04 & 3.10$e$-09 & 0 & 0 & 6.49$e$-06 & 1.56$e$-07 & 6.14$e$-06 & 2.56$e$-03 & 8.17$e$-03 & 0 & 2.39$e$-14 & 0 \\
\midrule
\multirow{2}{*}{11}             & bwd/ffn   & 3.05$e$-06 & 2.71$e$-11 & 5.02$e$-13 & 2.98$e$-11 & 7.21$e$-08 & 2.35$e$-09 & 4.24$e$-08 & 8.55$e$-05 & 1.06$e$-04 & 1.70$e$-10 & 1.04$e$-09 & 0 \\
              & bwd/attn  & 2.00$e$-04 & 4.44$e$-09 & 0 & 0 & 4.82$e$-06 & 1.21$e$-07 & 5.52$e$-06 & 2.30$e$-03 & 7.42$e$-03 & 0 & 7.18$e$-14 & 0 \\
\midrule
\multirow{2}{*}{10}             & bwd/ffn   & 3.02$e$-06 & 2.42$e$-11 & 3.35$e$-13 & 0 & 7.83$e$-08 & 2.34$e$-09 & 4.00$e$-08 & 8.64$e$-05 & 1.00$e$-04 & 3.44$e$-11 & 1.49$e$-09 & 0 \\
              & bwd/attn  & 1.65$e$-04 & 2.88$e$-09 & 0 & 5.40$e$-10 & 4.48$e$-06 & 1.05$e$-07 & 5.13$e$-06 & 2.30$e$-03 & 6.94$e$-03 & 0 & 2.39$e$-14 & 0 \\
\midrule
\multirow{2}{*}{9}             & bwd/ffn   & 2.97$e$-06 & 2.79$e$-11 & 3.01$e$-12 & 0 & 7.12$e$-08 & 2.32$e$-09 & 3.99$e$-08 & 8.70$e$-05 & 1.09$e$-04 & 2.41$e$-10 & 5.32$e$-10 & 0 \\
              & bwd/attn  & 1.77$e$-04 & 4.29$e$-09 & 0 & 0 & 4.65$e$-06 & 1.20$e$-07 & 5.41$e$-06 & 2.38$e$-03 & 7.29$e$-03 & 0 & 3.03$e$-10 & 0 \\
\midrule
\multirow{2}{*}{8}             & bwd/ffn   & 2.91$e$-06 & 2.29$e$-11 & 1.19$e$-13 & 0 & 7.08$e$-08 & 2.26$e$-09 & 4.05$e$-08 & 8.10$e$-05 & 1.12$e$-04 & 6.64$e$-11 & 3.18$e$-10 & 0 \\
              & bwd/attn  & 1.31$e$-04 & 4.38$e$-09 & 0 & 0 & 3.70$e$-06 & 8.33$e$-08 & 3.62$e$-06 & 1.91$e$-03 & 6.14$e$-03 & 0 & 4.79$e$-14 & 0 \\
\midrule
\multirow{2}{*}{7}             & bwd/ffn   & 2.83$e$-06 & 2.09$e$-11 & 2.70$e$-12 & 0 & 6.70$e$-08 & 2.21$e$-09 & 3.95$e$-08 & 8.51$e$-05 & 1.12$e$-04 & 1.04$e$-10 & 2.98$e$-09 & 0 \\
              & bwd/attn  & 9.79$e$-05 & 1.14$e$-09 & 0 & 1.96$e$-08 & 2.91$e$-06 & 6.57$e$-08 & 2.35$e$-06 & 1.77$e$-03 & 5.66$e$-03 & 0 & 2.39$e$-14 & 0 \\
\midrule
\multirow{2}{*}{6}             & bwd/ffn   & 2.73$e$-06 & 2.90$e$-11 & 2.44$e$-12 & 3.31$e$-11 & 6.51$e$-08 & 2.15$e$-09 & 3.92$e$-08 & 7.85$e$-05 & 1.13$e$-04 & 1.74$e$-10 & 3.99$e$-09 & 2.36$e$-14 \\
              & bwd/attn  & 1.05$e$-04 & 1.66$e$-09 & 0 & 0 & 2.98$e$-06 & 6.47$e$-08 & 3.14$e$-06 & 1.54$e$-03 & 5.96$e$-03 & 0 & 2.85$e$-10 & 0 \\
\midrule
\multirow{2}{*}{5}             & bwd/ffn   & 2.63$e$-06 & 2.75$e$-11 & 9.56$e$-14 & 0 & 5.98$e$-08 & 2.07$e$-09 & 3.72$e$-08 & 7.64$e$-05 & 1.06$e$-04 & 2.26$e$-10 & 8.34$e$-09 & 0 \\
              & bwd/attn  & 9.13$e$-05 & 3.35$e$-09 & 0 & 0 & 3.19$e$-06 & 5.74$e$-08 & 2.67$e$-06 & 1.29$e$-03 & 5.54$e$-03 & 0 & 1.73$e$-09 & 0 \\
\midrule
\multirow{2}{*}{4}             & bwd/ffn   & 2.54$e$-06 & 2.14$e$-11 & 7.17$e$-14 & 0 & 5.38$e$-08 & 2.02$e$-09 & 3.77$e$-08 & 7.44$e$-05 & 1.13$e$-04 & 8.00$e$-11 & 3.22$e$-09 & 0 \\
              & bwd/attn  & 7.84$e$-05 & 1.84$e$-09 & 0 & 1.27$e$-10 & 2.97$e$-06 & 5.82$e$-08 & 2.28$e$-06 & 1.05$e$-03 & 5.14$e$-03 & 0 & 1.70$e$-09 & 0 \\
\midrule
\multirow{2}{*}{3}             & bwd/ffn   & 2.43$e$-06 & 2.24$e$-11 & 7.17$e$-14 & 0 & 5.32$e$-08 & 1.97$e$-09 & 3.78$e$-08 & 7.65$e$-05 & 1.07$e$-04 & 3.70$e$-11 & 8.23$e$-09 & 0 \\
              & bwd/attn  & 6.95$e$-05 & 1.62$e$-09 & 2.39$e$-14 & 0 & 2.56$e$-06 & 4.71$e$-08 & 2.45$e$-06 & 1.05$e$-03 & 5.87$e$-03 & 0 & 2.24$e$-09 & 0 \\
\midrule
\multirow{2}{*}{2}             & bwd/ffn   & 2.28$e$-06 & 1.63$e$-11 & 0 & 0 & 4.31$e$-08 & 1.80$e$-09 & 3.36$e$-08 & 6.95$e$-05 & 1.02$e$-04 & 7.85$e$-12 & 2.47$e$-09 & 0 \\
              & bwd/attn  & 8.15$e$-05 & 1.15$e$-09 & 0 & 3.03$e$-09 & 3.86$e$-06 & 6.08$e$-08 & 3.32$e$-06 & 1.06$e$-03 & 7.60$e$-03 & 0 & 1.27$e$-09 & 0 \\
\midrule
\multirow{2}{*}{1}             & bwd/ffn   & 2.12$e$-06 & 1.14$e$-11 & 1.34$e$-12 & 3.70$e$-10 & 8.74$e$-08 & 1.62$e$-09 & 3.03$e$-08 & 6.46$e$-05 & 1.33$e$-04 & 6.57$e$-11 & 9.29$e$-09 & 0 \\
              & bwd/attn  & 1.03$e$-04 & 2.30$e$-09 & 0 & 0 & 5.05$e$-06 & 5.76$e$-08 & 2.10$e$-06 & 1.07$e$-03 & 4.40$e$-03 & 0 & 2.39$e$-14 & 0 \\
\bottomrule
\end{tabular}

\end{sc}
\end{tiny}
\end{center}
\vskip 0.15in

\caption{Frequency of mismatching tensor elements for each transformer primitive backward, separated down by decoder layer and averaged across all microsteps. The unhealthy Nodes 2, 3, 5, 12, and 13 did not exhibit any mismatching tensors in backward passes in this experimental setting and thus are excluded from the table.}

\label{fig:backwards_primitive_frequency}
\end{table*}

\begin{table*}[t]

\begin{center}
\begin{tiny}
\begin{sc}

\begin{tabular}{rrrrrrrrrrr}
\toprule
 &  & \multicolumn{9}{c}{SDC Node ID}                                                \\
\midrule
Decoder Layer        & Primitive            & Node 1     & Node 4      & Node 6      & Node 7       & Node 8       & Node 9          & Node 10      & Node 11     & Node 14     \\
\midrule
\multirow{2}{*}{1}                    & fwd/attn             & 1.82  & 0.3047 & 0      & 4.7812  & 13.5    & 0.15039063 & 0       & 25.875 & 0      \\
                     & fwd/ffn              & 63.25 & 0.0512 & 0      & 16.6250 & 0.3867  & 21.625     & 2.6875  & 442    & 0      \\
\midrule
\multirow{2}{*}{2}                     & fwd/attn             & 1.04  & 0.3515 & 0.0366 & 24.625  & 0.1289  & 1.4140625  & 4.84375 & 90.5   & 0.4746 \\
                     & fwd/ffn              & 312   & 0.0266 & 1      & 1.3046  & 0.0981  & 60.25      & 80      & 668    & 0      \\
\midrule
\multirow{2}{*}{3}                   & fwd/attn             & 3.97  & 2.9531 & 0      & 6.125   & 2.7031  & 6.4375     & 12.9375 & 230    & 0      \\
                     & fwd/ffn              & 84.5  & 0.0439 & 0      & 9.8125  & 0.2793  & 15.5625    & 42.5    & 848    & 0      \\
\midrule
\multirow{2}{*}{4}                    & fwd/attn             & 4.91  & 0.375  & 0      & 6.8438  & 0.7148  & 4.59375    & 24.25   & 188    & 1.1172 \\
                     & fwd/ffn              & 92    & 0.8555 & 0      & 1.7734  & 0.7734  & 130        & 145     & 640    & 0      \\
\midrule
\multirow{2}{*}{5}                  & fwd/attn             & 11.63 & 0.1699 & 0      & 9.375   & 0.6406  & 7.75       & 1120    & 127    & 0.3613 \\
                     & fwd/ffn              & 18.38 & 0.0556 & 0      & 17.125  & 0.1338  & 30.75      & 118     & 708    & 0      \\
\midrule
\multirow{2}{*}{6}                    & fwd/attn             & 99    & 0.3281 & 0      & 11.9375 & 4.2187  & 16.5       & 58.75   & 152    & 0      \\
                     & fwd/ffn              & 143   & 0.0737 & 0.0933 & 25      & 4.0312  & 24.75      & 67.5    & 298    & 0      \\
\midrule
\multirow{2}{*}{7}                    & fwd/attn             & 14.63 & 0.1069 & 0.3847 & 43.25   & 0.7695  & 12.9375    & 74.5    & 76.5   & 0      \\
                     & fwd/ffn              & 88.5  & 0.0181 & 0      & 2.4219  & 0.1719  & 200        & 40.5    & 334    & 0      \\
\midrule
\multirow{2}{*}{8}                     & fwd/attn             & 12.94 & 0.0933 & 0      & 6.9375  & 0.5898  & 7.15625    & 36.75   & 76     & 0      \\
                     & fwd/ffn              & 145   & 0.0913 & 0      & 6.125   & 4.375   & 97         & 47.25   & 422    & 0      \\
\midrule
\multirow{2}{*}{9}                    & fwd/attn             & 38.5  & 0.7852 & 0.9961 & 1.9063  & 1.1016  & 18.25      & 67      & 318    & 0.0786 \\
                     & fwd/ffn              & 148   & 0.0579 & 0      & 25.625  & 5.6875  & 21.875     & 53      & 214    & 0      \\
\midrule
\multirow{2}{*}{10}                    & fwd/attn             & 9.81  & 1.125  & 0.1289 & 4.5938  & 4.8125  & 6.46875    & 42.75   & 136    & 0      \\
                     & fwd/ffn              & 119.5 & 0.0287 & 0      & 28.375  & 0.1533  & 30.625     & 107.5   & 398    & 0      \\
\midrule
\multirow{2}{*}{11}                    & fwd/attn             & 11.69 & 0.1650 & 0      & 6.0625  & 0.6015  & 14.3125    & 53      & 121    & 0      \\
                     & fwd/ffn              & 33.75 & 0.0425 & 0.1064 & 13.8125 & 1.3437  & 35         & 76.5    & 346    & 0      \\
\midrule
\multirow{2}{*}{12}                    & fwd/attn             & 25.5  & 0.2832 & 0      & 25.75   & 0.4160  & 3.21875    & 27.625  & 34     & 0      \\
                     & fwd/ffn              & 23.25 & 0.0349 & 1      & 88.5    & 0.4394  & 139        & 262     & 296    & 0      \\
\midrule
\multirow{2}{*}{13}                   & fwd/attn             & 1.36  & 0.4648 & 0.3632 & 3.4843  & 0.0054  & 4.78125    & 43.75   & 48.5   & 0.0198 \\
                     & fwd/ffn              & 37.75 & 1.4609 & 0      & 7.8437  & 0.3105  & 18.75      & 80.5    & 976    & 0      \\
\midrule
\multirow{2}{*}{14}                    & fwd/attn             & 2.58  & 0.4102 & 1.0078 & 7.25    & 0.2617  & 4.96875    & 93.5    & 28.5   & 0      \\
                     & fwd/ffn              & 36.5  & 0.0232 & 0.9961 & 5.8125  & 0.21875 & 79.5       & 73.5    & 432    & 0      \\
\midrule
\multirow{2}{*}{15}                    & fwd/attn             & 2.25  & 0.0928 & 0      & 7.2813  & 0.1445  & 12.5625    & 47.25   & 49.25  & 0      \\
                     & fwd/ffn              & 97.5  & 0.0322 & 0      & 1.6953  & 0.5     & 23.375     & 31.625  & 366    & 0      \\
\midrule
\multirow{2}{*}{16}                    & fwd/attn             & 11.94 & 0.9023 & 1      & 4.625   & 0.0971  & 119        & 40.75   & 103    & 0.4707 \\
                     & fwd/ffn              & 39.75 & 0.0273 & 0      & 1.6641  & 0.1035  & 115        & 45.75   & 536    & 0     \\
\bottomrule
\end{tabular}

\end{sc}
\end{tiny}
\end{center}
\vskip 0.15in

\caption{Severity of SDCs in transformer primitive forwards, separated down by decoder layer and maximized over all microsteps. The unhealthy Nodes 2, 3, 5, 12, and 13 and 15 did not exhibit any mismatching tensors in forward passes in this experimental setting and thus are excluded from the table.}

\label{fig:forward_primitive_max_severity}
\end{table*}
\begin{table*}[t]

\begin{center}
\begin{tiny}
\begin{sc}

\setlength{\tabcolsep}{2pt}
\begin{tabular}{rrrrrrrrrrrrrr}
\toprule
 & & \multicolumn{12}{c}{SDC Node ID} \\
 \midrule
 Decoder Layer & Primitive & Node 1 & Node 4 & Node 5 & Node 6 & Node 7 & Node 8 & Node 9 & Node 10 & Node 11 & Node 13 & Node 14 & Node 15 \\
 \midrule
\multirow{2}{*}{16} & bwd/ffn & 0.7383 & 0.0942 & 0.0376 & 0 & 0.1367 & 0.1143 & 9.621$e$+12 & 0.1436 & 236 & 0.0588 & 0.0757 & 0.0053 \\
 & bwd/attn & 1256 & 0.3223 & 0 & 0 & 2.0938 & 3.3281 & 3.758$e$+11 & 1.1094 & 13.1875 & 0 & 0.0344 & 0 \\
\midrule
\multirow{2}{*}{15} & bwd/ffn & 0.9023 & 0.2598 & 0.0206 & 0 & 0.1406 & 0.7266 & 8.934$e$+12 & 0.2129 & 127 & 0.2344 & 0.0564 & 0 \\
 & bwd/attn & 968 & 0.1885 & 0 & 0 & 1.3516 & 0.4648 & 3.737$e$+11 & 0.3828 & 8.4375 & 0 & 0 & 0 \\
\midrule
\multirow{2}{*}{14} & bwd/ffn & 1.5156 & 0.3359 & 0.0223 & 0.0267 & 0.0972 & 0.8008 & 6.769$e$+12 & 0.4805 & 692 & 0.3164 & 0.1245 & 0 \\
 & bwd/attn & 1040 & 1.1641 & 0 & 0 & 1.1563 & 6.4063 & 1.154$e$+11 & 0.3125 & 39.75 & 0 & 0.0165 & 0 \\
\midrule
\multirow{2}{*}{13}& bwd/ffn & 3.8750 & 0.5117 & 0.0679 & 0.0371 & 0.2676 & 0.1338 & 8.212$e$+12 & 0.1563 & 352 & 0.0674 & 0.0596 & 0.0176 \\
 & bwd/attn & 7392 & 2.0313 & 0 & 0.0469 & 0.7070 & 0.1318 & 1.943$e$+11 & 0.3145 & 9.3125 & 0 & 0.0791 & 0 \\
\midrule
\multirow{2}{*}{12} & bwd/ffn & 2.7656 & 0.3398 & 0.0908 & 0.0309 & 0.2969 & 0.1914 & 5.429$e$+12 & 0.2559 & 1800 & 0.1768 & 0.0608 & 0 \\
 & bwd/attn & 1640 & 0.3633 & 0 & 0 & 0.9961 & 4.0000 & 2.212$e$+11 & 0.3379 & 68.5 & 0 & 0.0075 & 0 \\
\midrule
\multirow{2}{*}{11} & bwd/ffn & 0.6641 & 0.4434 & 0.0310 & 0.0525 & 0.2061 & 0.1250 & 8.624$e$+12 & 0.1631 & 7680 & 0.1299 & 0.0781 & 0 \\
 & bwd/attn & 1992 & 0.2852 & 0 & 0 & 1.3516 & 0.1143 & 1.750$e$+11 & 0.5625 & 2208 & 0 & 0.0952 & 0 \\
\midrule
\multirow{2}{*}{10} & bwd/ffn & 0.7852 & 1.0078 & 0.0078 & 0 & 0.1709 & 1.3359 & 5.601$e$+12 & 0.1924 & 332 & 0.0571 & 0.1494 & 0 \\
 & bwd/attn & 3488 & 1.5000 & 0 & 0.0625 & 1.6406 & 0.2793 & 1.299$e$+11 & 0.8828 & 55.25 & 0 & 0.0272 & 0 \\
\midrule
\multirow{2}{*}{9} & bwd/ffn & 0.6914 & 0.1025 & 0.0209 & 0 & 0.0977 & 1.5938 & 4.948$e$+12 & 0.3184 & 1776 & 0.0449 & 0.0562 & 0 \\
 & bwd/attn & 4672 & 1.1484 & 0 & 0 & 0.8086 & 0.6719 & 2.835$e$+11 & 0.2539 & 77.5 & 0 & 0.0239 & 0 \\
\midrule
\multirow{2}{*}{8} & bwd/ffn & 0.6445 & 0.1289 & 0.0164 & 0 & 0.1196 & 0.1816 & 3.522$e$+12 & 0.6367 & 256 & 0.0718 & 0.1367 & 0 \\
 & bwd/attn & 864 & 1.8359 & 0 & 0 & 1.1172 & 1.5469 & 9.073$e$+10 & 0.7617 & 29.5 & 0 & 0.0124 & 0 \\
\midrule
\multirow{2}{*}{7} & bwd/ffn & 0.6406 & 0.2539 & 0.0679 & 0 & 0.1021 & 0.1040 & 4.364$e$+12 & 0.2119 & 1880 & 0.1045 & 0.1465 & 0 \\
 & bwd/attn & 748 & 0.3027 & 0 & 0.1621 & 7.6563 & 0.3359 & 7.999$e$+10 & 4.0000 & 104.5 & 0 & 0.0266 & 0 \\
\midrule
\multirow{2}{*}{6} & bwd/ffn & 0.4434 & 0.1826 & 0.0320 & 0.0322 & 0.0835 & 0.5000 & 2.749$e$+12 & 0.2656 & 173 & 0.1172 & 0.1084 & 0.0082 \\
 & bwd/attn & 3248 & 0.8242 & 0 & 0 & 5.3438 & 0.9102 & 7.892$e$+10 & 12.7500 & 76 & 0 & 0.0223 & 0 \\
\midrule
\multirow{2}{*}{5} & bwd/ffn & 0.3320 & 0.6172 & 0.0080 & 0 & 0.1533 & 0.1040 & 2.233$e$+12 & 0.6484 & 1016 & 0.0742 & 0.3789 & 0 \\
 & bwd/attn & 664 & 0.5469 & 0 & 0 & 6.4063 & 0.6875 & 7.087$e$+10 & 0.9492 & 190 & 0 & 0.3828 & 0 \\
\midrule
\multirow{2}{*}{4} & bwd/ffn & 0.4883 & 0.2852 & 0.0082 & 0 & 0.1074 & 0.1738 & 1.563$e$+12 & 0.2246 & 334 & 0.0664 & 0.0781 & 0 \\
 & bwd/attn & 1784 & 0.1816 & 0 & 0.0356 & 3.6094 & 0.1895 & 8.536$e$+10 & 3.3750 & 304 & 0 & 3.7969 & 0 \\
\midrule
\multirow{2}{*}{3} & bwd/ffn & 0.4043 & 5.3750 & 0.0093 & 0 & 0.1729 & 0.1064 & 1.623$e$+12 & 0.2178 & 556 & 0.0879 & 0.2295 & 0 \\
 & bwd/attn & 1020 & 2.3906 & 0.0048 & 0 & 1.6172 & 0.5000 & 9.342$e$+10 & 1.0859 & 74 & 0 & 1.0156 & 0 \\
\midrule
\multirow{2}{*}{2} & bwd/ffn & 1.6328 & 0.8008 & 0 & 0 & 0.1387 & 0.0796 & 1.512$e$+12 & 0.1416 & 836 & 0.1157 & 0.1128 & 0 \\
 & bwd/attn & 644 & 0.3594 & 0 & 0.0537 & 0.9844 & 0.2080 & 8.966$e$+10 & 1.3359 & 216 & 0 & 2.2500 & 0 \\
\midrule
\multirow{2}{*}{1} & bwd/ffn & 0.6680 & 0.1152 & 0.0195 & 0.0776 & 0.1641 & 0.0801 & 1.044$e$+12 & 0.2344 & 406 & 0.0356 & 0.1270 & 0 \\
 & bwd/attn & 596 & 0.3535 & 0 & 0 & 32.2500 & 0.1436 & 1.463$e$+10 & 0.2676 & 241 & 0 & 0.2656 & 0 \\
 \bottomrule
\end{tabular}

\end{sc}
\end{tiny}
\end{center}
\vskip 0.15in

\caption{Severity of SDCs in transformer primitive backwards, separated down by decoder layer and maximized over all microsteps. The unhealthy Nodes 2, 3, and 12 did not exhibit any mismatching tensors in forward passes in this experimental setting and thus are excluded from the table.}

\label{fig:backwards_primitive_max_severity}
\end{table*}

\end{document}